\PassOptionsToPackage{table}{xcolor}
\RequirePackage{fix-cm}
\documentclass{article}
\usepackage[OT1]{fontenc}
\usepackage{spatialmind,times}

\usepackage{amsmath,amsfonts}
\usepackage{url}
\usepackage{booktabs}
\usepackage{float,placeins}
\usepackage[table]{xcolor}
\usepackage{array,tabularx}
\usepackage{graphicx}
\usepackage{microtype}
\usepackage{hyperref}

\definecolor{SMHeader}{HTML}{F1F4F7}
\definecolor{SMGroup}{HTML}{EAF2F8}
\definecolor{SMOurs}{HTML}{DBEAF5}
\definecolor{SMAccent}{HTML}{2B557A}
\definecolor{SMMuted}{HTML}{596774}
\definecolor{SMReference}{HTML}{F6F7F8}

\newcolumntype{Y}{>{\raggedleft\arraybackslash}X}
\newcolumntype{C}{>{\centering\arraybackslash}X}

\newcommand{\SMbest}[1]{\textbf{#1}}
\newcommand{\SMgroup}[1]{\textcolor{SMAccent}{\textit{#1}}}

\title{From Sight to Foresight: Predictive Spatial Reasoning in Vision-Language Models}

\author{
    \href{https://openreview.net/profile?id=~Feiran_Wang2}{Feiran Wang}$^{1}$\thanks{\protect\hypertarget{author-notes}{}This work was done during an internship at Bosch.\quad $^{\ddagger}$Project lead.},
    \href{https://openreview.net/profile?id=~Xiaoqi_Wang2}{Xiaoqi Wang}$^{2,3}$\hyperlink{author-notes}{$^{\ddagger}$},
    \href{https://openreview.net/profile?id=~Ziwei_Li3}{Ziwei Li}$^{2,3}$,
    \href{https://openreview.net/profile?id=~Wenbin_He1}{Wenbin He}$^{2,3}$,
    \href{https://openreview.net/profile?id=~Yan_Yan6}{Yan Yan}$^{1}$,
    \href{https://openreview.net/profile?id=~Liu_Ren1}{Liu Ren}$^{2,3}$ \\
    {\normalfont $^{1}$University of Illinois at Chicago,} \\
    {\normalfont $^{2}$Bosch Research North America, $^{3}$Bosch Center for Artificial Intelligence (BCAI)}
}

\spatialmindfinalcopy
\hypersetup{
    hidelinks,
    pdftitle={From Sight to Foresight: Predictive Spatial Reasoning in Vision-Language Models},
    pdfauthor={Feiran Wang, Xiaoqi Wang, Ziwei Li, Wenbin He, Yan Yan, Liu Ren}
}

\newcommand{\Eqref}[1]{Equation~\ref{#1}}

\let\cite\citep
\begin{document}
\raggedbottom
\microtypesetup{activate=false}
\setcounter{footnote}{1}
\maketitle
\begingroup
\setlength{\intextsep}{0pt}
\begin{figure}[htbp]
    \vspace{-0.2em}
    \centering
    \includegraphics[width=\linewidth, trim={4pt 4pt 4pt 4pt}]{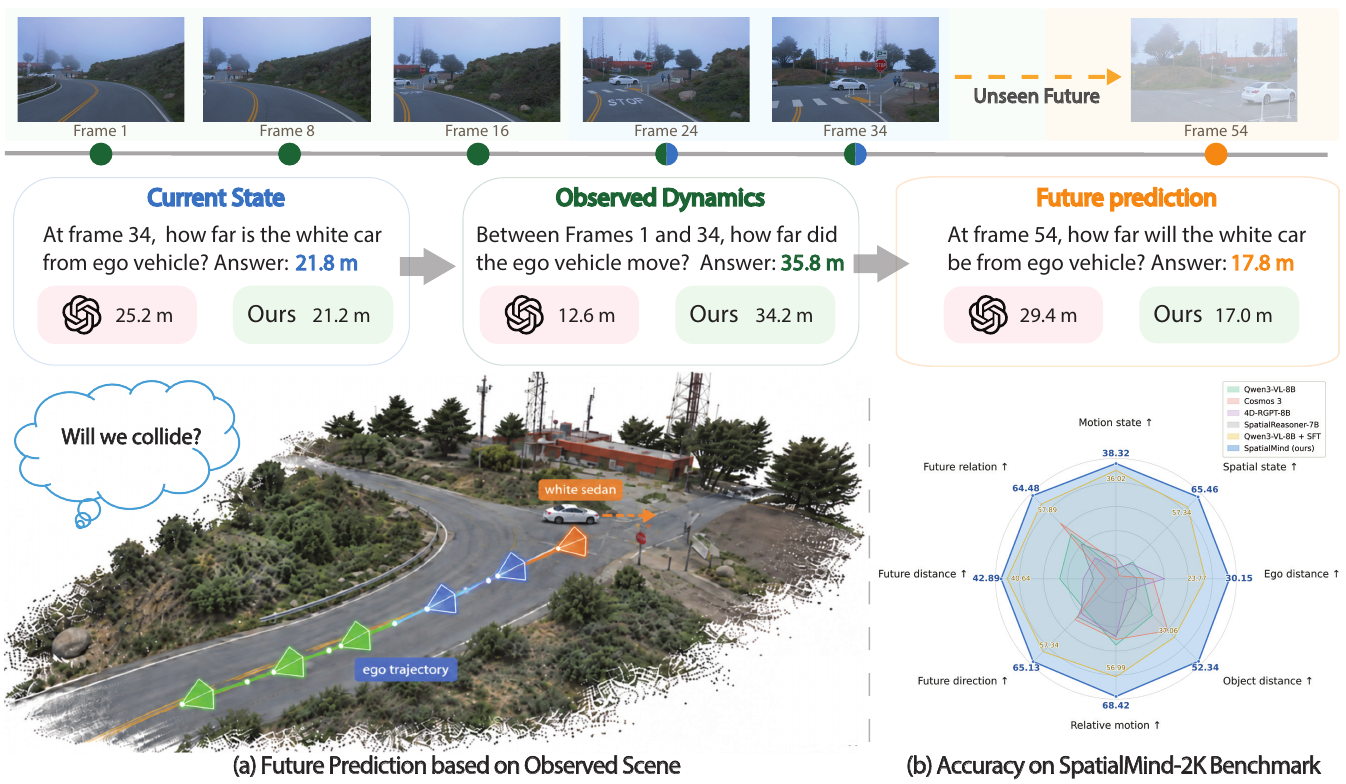}
    \vspace{-1.5em}
    \caption{
    SpatialMind enables future spatial prediction by reasoning progressively from observed evidence, which can help identify collision risks and support timely action to avoid hazards.  By grounding prediction in current geometry and observed dynamics, SpatialMind yields more accurate estimates than a proprietary VLM. (a) The 3D scene illustrates the spatial context for anticipating future interactions between the ego vehicle and the white sedan. (b) Benchmark results show consistent improvements over the compared models across all eight spatial reasoning tasks.
    }
    \label{fig:teaser}
\end{figure}

\endgroup
\vspace{12pt}
\begin{abstract}
    \vspace{-1em}
    Predicting future spatial states supports collision avoidance and timely
    decision-making in dynamic environments.
    However, existing vision-language models (VLMs) and benchmarks for spatial
    reasoning primarily focus on observed scenes, leaving predictive spatial
    reasoning beyond the observed interval underexplored.
    To this end, we introduce SpatialMind, a metric-scale VLM for spatial
    reasoning and future prediction. Its metric depth adapter anchors spatial
    reasoning to real-world scale, while its progressive state chain establishes
    current spatial states and observed dynamics as the foundation for future
    prediction.
    Given a video prefix, SpatialMind predicts distances, motion directions,
    and spatial relations in both observed and unseen future frames.
    For training and evaluation, we build a scalable data engine that grounds
    entity descriptions in metric geometry to generate question-answer pairs
    and state supervision. Using this engine, we construct the SpatialMind-30K
    dataset and the SpatialMind-2K benchmark, both covering driving and everyday
    egocentric scenes. The benchmark spans eight tasks across three levels:
    current-state understanding, observed-dynamics understanding, and future
    prediction.
    Experiments show that SpatialMind substantially outperforms both general
    and spatially specialized models on our benchmark while achieving
    competitive zero-shot performance on VSI-Bench, OSI-Bench, and VLM4D.
    The project page and code are available at {\urlstyle{same}\url{https://brack-wang.github.io/spatialmind/}}.
\end{abstract}

\section{Introduction}
\label{sec:introduction}

Humans anticipate future changes in distances and spatial relations, enabling navigation and interaction with the physical world. In dynamic environments, future prediction is important for assessing whether an action can be completed safely. In Figure~\ref{fig:teaser}, a driver can instinctively anticipate the narrowing gap to the white car and assess whether sufficient clearance will remain during braking. Cognitive studies associate this capacity with \emph{prospection}:
using observed experience to reason beyond immediate perception and anticipate future changes~\citep{craik1967nature,BUCKNER200749,schacter2007remembering}. While current VLMs have made substantial progress in understanding observed visual content, extending this understanding to physically grounded future spatial prediction remains challenging~\citep{ishihara2026stride,zhou2025vlm4d,fan2026vlm}. This raises a fundamental question: what spatial understanding does a VLM need for reliable future prediction?

Reliable future prediction depends on understanding the current scene and its observed evolution. To understand the current scene, a model first needs to ground entities in metric space and establish their spatial relations, which anchors future prediction in metric geometry. To determine how the scene has changed over the observed interval, it then needs to estimate motion and track changes in spatial relations, which reveals temporal trends that indicate how the scene may evolve. To support VLMs in extending their spatial reasoning from understanding observations to predicting future states, we organize spatial reasoning into three progressive levels: (i) \emph{current state understanding}, which establishes metric geometry and spatial relations; (ii) \emph{observed dynamics understanding}, which captures how the scene evolves throughout the observed interval; and (iii) \emph{future prediction}, which extrapolates this evolution to anticipate subsequent spatial changes.

Existing VLMs and spatial datasets cover only parts of this progression, with future prediction receiving far less attention than reasoning over observed scenes. Recent models have advanced spatial reasoning~\citep{cheng2024spatialrgpt,cheng20263d,fan2026vlm,yang20264d,yin2026mllm4dvisualbasedspatialtemporalintelligence}, yet many focus on static geometry, explicitly prompted regions, or dynamics within the observed sequence, leaving prediction beyond the observations less explored. Training datasets that provide consistent supervision across the three levels for the same entities also remain limited. Existing benchmarks cover complementary aspects of metric geometry, observed motion, and short-horizon prediction~\citep{li2025sti,zhou2025vlm4d,wu2026indoor,ishihara2026stride,jia2026omnispatial}, but provide limited evaluation of the full progression from current-scene understanding through observed dynamics to future prediction. It therefore remains unclear whether models can use their understanding of current states and observed changes to anticipate future spatial conditions.

To enable and systematically evaluate future prediction
in VLMs,
we introduce \textbf{SpatialMind}, a spatial reasoning VLM designed to
build and update a coherent spatial mind state from observation to prediction.
To ground spatial reasoning at real-world scale, SpatialMind augments visual
representations with metric depth and motion cues from a pretrained spatial
expert.
To model the spatial changes, we introduce a \emph{progressive state chain} to structure the model's reasoning output across the three levels: grounding the referred entity and establishing its current spatial state, updating this state from observed dynamics, and extrapolating to the near future when required.
This design
connects metric grounding, temporal updating, and future reasoning within a
single evolving state.
To support training and evaluation, we develop a scalable data engine that grounds entity descriptions in metric geometry to generate QA pairs from both driving and everyday egocentric scenes.  Using this engine, we construct SpatialMind-30K Dataset with 30,677 training QA pairs and SpatialMind-2K Benchmark with 2,000 evaluation questions. Both cover eight question categories across the same three levels: metric spatial states, observed ego and object dynamics, and future spatial relations.

Experiments on SpatialMind-2K Benchmark reveal persistent limitations in metric reasoning and future prediction among existing VLMs. SpatialMind substantially outperforms these models across all three reasoning levels, with consistent gains over the fine-tuned baseline model on the same training data. Our competitive zero-shot performance on VSI-Bench, OSI-Bench, and VLM4D further demonstrates its robust generalization capability across diverse spatial reasoning tasks.

Our main contributions are summarized as follows:

\begin{itemize}
    \item We introduce SpatialMind, a method that significantly enhances metric distance estimation and spatiotemporal relation understanding, using observed dynamics to infer future spatial states and showing strong spatial reasoning and future prediction capabilities.
    \item We build a scalable data curation pipeline that generates geometrically grounded spatiotemporal QA pairs from both driving and everyday scenes, totaling 30,677 QA pairs with corresponding reasoning chains. We further introduce SpatialMind-2K Benchmark to evaluate VLM capabilities in metric-scale estimation, temporal reasoning, and future prediction.
    \item Experiments demonstrate that our SpatialMind model gains substantial improvement on SpatialMind-2K and competitive performance on VSI-Bench, OSI-Bench, and VLM4D.
\end{itemize}

\section{Related Work}

\textbf{Spatial Reasoning in VLMs.}
Existing approaches to spatial reasoning in VLMs mainly incorporate external geometric priors through depth supervision, RGB-D inputs, or pretrained 3D features~\citep{chen2024spatialvlm,cai2025spatialbot,cheng2024spatialrgpt,fan2026vlm,zheng2026learning}.
Although these representations provide explicit spatial cues, their metric precision remains constrained by the scale accuracy of the underlying geometric estimates.
Complementary work enables future prediction through future-event supervision, temporal event chains, or latent visual reasoning~\citep{wang2026fostering,su2026video,jiang2026imagine}.
These methods primarily target future events, with less emphasis on explicitly estimating spatial quantities such as future distances in meters.
SpatialMind integrates an absolute metric adapter with a progressive state chain to support spatial reasoning from current state, observed dynamics to future spatial prediction at real-world scale.

\textbf{Spatial Reasoning Datasets and Benchmarks.}
Spatial reasoning benchmarks have expanded from static spatial relations to metric and spatiotemporal understanding~\citep{ma20253dsrbench,jia2026omnispatial,yang2026mmsi,wu2026spatialscore}.
Existing evaluations emphasize spatial measurements, viewpoint-dependent relations, and observed motion~\citep{yang2025thinking,zhang2026flatland,li2025sti,zhou2025vlm4d,fan2026vlm,wu2026indoor}.
These benchmarks mainly assess geometry and dynamics within observed scenes, with limited coverage of spatial states beyond the observed interval.
OmniSpatial and SCP-Bench extend evaluation to motion prediction and inference about unseen spatial states, respectively~\citep{jia2026omnispatial,zhao2026scp}.
STRIDE-QA~\citep{ishihara2026stride} further evaluates numerical predictions of future distances and velocities, but focuses on short-horizon prediction in urban driving.
However, future spatial prediction at physical scale lacks systematic evaluation and training data. To fill this gap, SpatialMind offers a dataset and benchmark spanning eight question categories, from observed dynamics to future prediction.

\section{SpatialMind Data Engine}
\label{sec:data_engine}

SpatialMind supports metric-scale spatial reasoning and future prediction from video. Our data engine constructs metric geometry, establishes entity correspondences, and generates QA pairs with spatial metadata. Using this engine, we build a dataset and benchmark covering three reasoning levels and eight tasks. We further introduce the SpatialMind model in Sec.~\ref{sec:model}, which we train on this dataset and evaluate on the benchmark to assess its predictive spatial reasoning capabilities.

\subsection{Target Selection and Unique Description}
\label{sec:linguistic_grounding}
\begin{figure}[t]
    \centering
    \includegraphics[width=1.0\linewidth]{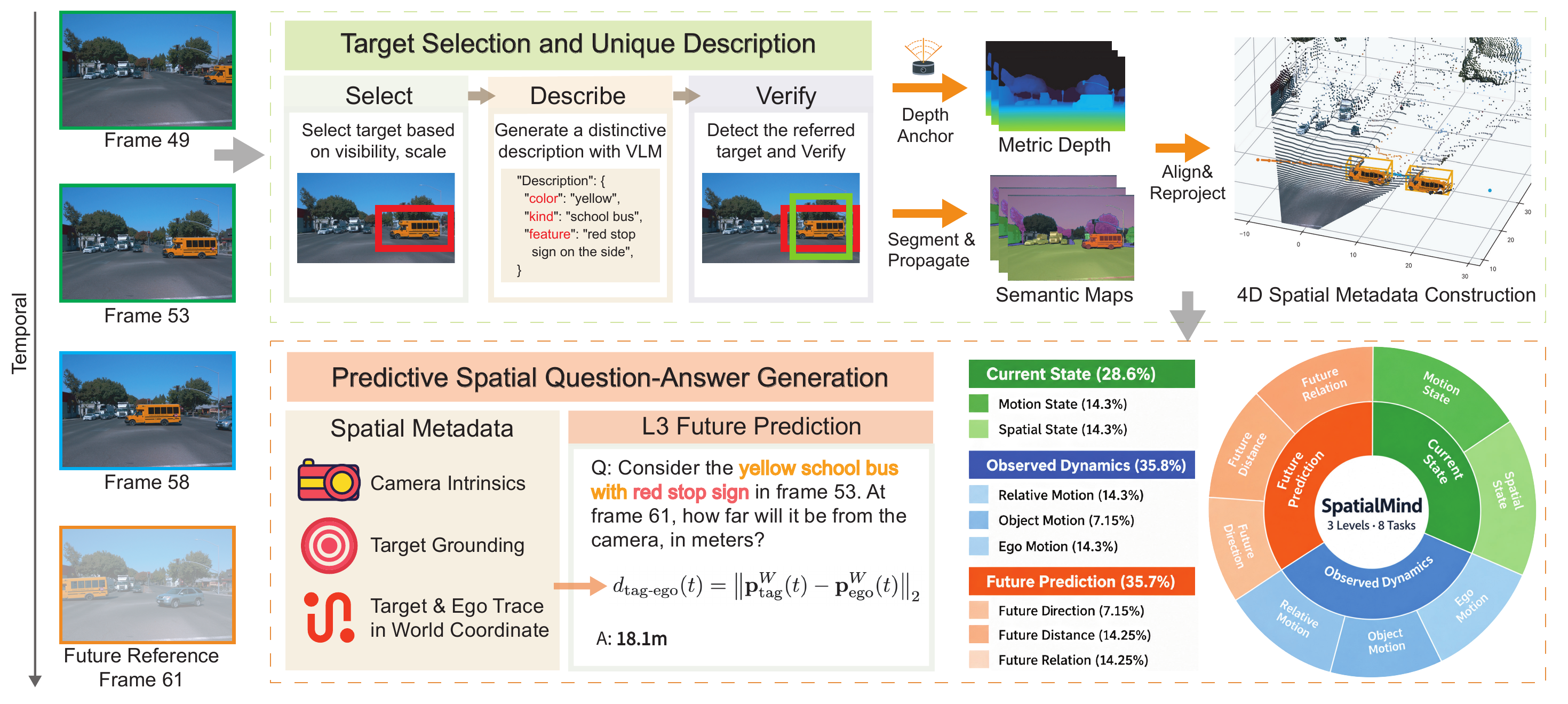}
    \vspace{-2em}
    \caption{
    \textbf{SpatialMind Data Engine.} We assign each target a distinctive natural-language description through a Select--Describe--Verify process, then combine metric geometry, semantic propagation, and temporal alignment to construct spatial metadata, which supports automatic QA generation across three levels and eight spatial reasoning tasks, with the data distribution shown on the right.
    }
    \vspace{-1em}

    \label{fig:data}
\end{figure}

Constructing spatial questions involves selecting target objects and providing descriptions that identify each one uniquely. To this end, we follow three steps: target selection, description generation, and verification (Figure~\ref{fig:data}). To track objects over time, we apply a video instance segmentation model that segments individual objects and associates their regions across frames. To keep appearance details recognizable, we select targets that remain visible across consecutive frames and occupy sufficient image area. To capture the richest visual detail for each target, we take the frame in which it occupies the largest image area as the reference frame, crop the target region, and send the crop to a VLM, which describes the object's category, appearance, and distinctive attributes.

To verify that each description identifies its intended target uniquely, we perform reverse localization. Given the description and an unmarked frame from the original video, a VLM predicts a bounding box for the described object and checks whether any other object also matches the description. We retain a description only when the predicted box sufficiently overlaps the target's box and the VLM finds no other matching object.
Together, the selected targets and their verified descriptions form the semantic basis for constructing spatial metadata in the next stage.

\subsection{4D Spatial Metadata Construction}
\label{sec:metric_annotation}

Constructing spatial reasoning data depends on scene geometry at real-world scale. Sparse depth measurements from source data provide physical scale but have limited coverage on individual targets, which can introduce errors into the spatial measurements used for Question-Answer pairs generation. To obtain dense depth while preserving physical scale, we first predict dense depth with a feed-forward 3D model. We then use the sparse source measurements as metric references to estimate a scale and offset, calibrating the predicted depth to real-world scale.  The calibrated depth provides dense geometry of visible surfaces, from which we measure entity positions and distances. To cover diverse scenes and spatial scales, we apply this procedure to outdoor driving videos from Waymo~\citep{sun2020scalability} and indoor egocentric videos from Ego-Exo4D~\citep{grauman2024ego}.

Predictive spatial reasoning questions need the states of targets and the observer over time, both for generating the questions and computing their reference answers. To localize each target from Sec.~\ref{sec:linguistic_grounding} in 3D, we unproject the calibrated depth within its mask into a shared world coordinate system using the camera intrinsics and poses. To recover each target's 4D trajectory, we propagate its mask across the video and link the resulting 3D localizations over time. To measure relative directions, we derive the observer's trajectory from the camera poses, taking the camera center at each frame as the observer's position and the camera orientation as its heading. The resulting spatial metadata comprises the targets' 4D trajectories, the observer's trajectory, and the verified referring descriptions that uniquely identify each target. This metadata serves as the computational basis for constructing spatial reasoning and future prediction questions.

\subsection{Predictive Spatial QA Generation}
\label{sec:qa_generation}

We generate spatial QA pairs for eight tasks across three progressive levels: current-state understanding, observed-dynamics understanding, and future prediction. Current states describe entity positions and local motion, while observed changes provide evidence for future prediction. We compute reference answers from spatial metadata using each entity description, spatial reference, and temporal scope. Mathematically, we use $t_s$ and $t_o$ to denote the start and end of the observed interval, and $\mathbf p^W(t)$ to denote an entity's position in world coordinates at frame $t$.

\noindent\textbf{Level~1: Current-state understanding} describes where entities are and how they are moving at a given time. We design \emph{spatial-state} questions about distances between entities and \emph{motion-state} questions about speed and turning angle. At frame $t$, we compute the target--observer distance as
\begin{equation}
    d(t)=\left\|\mathbf p_{\mathrm{tag}}^W(t)-\mathbf p_{\mathrm{ego}}^W(t)\right\|_2, \qquad t_s\leq t\leq t_o.
    \label{eq:current_spatial_state}
\end{equation}

Here, $d(t)$ denotes the target--observer distance in meters, and $\|\cdot\|_2$ denotes the Euclidean norm. These questions jointly assess the model's understanding of current spatial relations and motion, which provides a foundation for reasoning about observed dynamics and future prediction.

\noindent\textbf{Level~2: Observed-dynamics understanding} measures how far entities travel and how spatial relations change over an observed interval. We design \emph{ego-motion distance}, \emph{target-motion distance}, and \emph{relative-motion} questions. For the observer or target, the total distance traveled over $[t_s,t_o]$ is
\begin{equation}
    s_u(t_s,t_o)=\sum_{t=t_s}^{t_o-1}\left\|\mathbf p_u^W(t+1)-\mathbf p_u^W(t)\right\|_2, \qquad u\in\{\mathrm{ego},\mathrm{tag}\}.
    \label{eq:observed_dynamics}
    \vspace{-0.5em}
\end{equation}

Here, $s_u(t_s,t_o)$ denotes the path length of entity $u$ over the observed interval, measured in meters. Ego-motion distance, target-motion distance, and relative-motion questions jointly assess the model's ability to track entity movement and changing spatial relations across frames while accounting for observer motion. Observed dynamics provide motion evidence for future spatial prediction. Appendix~\ref{app:visualizations} illustrates a Level~2 relative-motion example, linking the observed video, metric geometry, and progressive state-chain supervision.

\noindent\textbf{Level~3: Future prediction} asks where entities will be and how entities will move beyond the observed interval. We split each source video into an observed prefix over $[t_s,t_o]$ and a withheld future segment. Using only the prefix as visual input, the model answers \emph{ego future direction}, \emph{target future distance}, and \emph{target future relation} questions about the future segment. We compute reference answers from annotations of the future segment; for a distance query at $t_f$, the reference answer is
\begin{equation}
    d(t_f)=\left\|\mathbf p_{\mathrm{tag}}^W(t_f)-\mathbf p_{\mathrm{ego}}^W(t_f)\right\|_2, \qquad t_f>t_o.
    \label{eq:future_spatial_prediction}
\end{equation}

Here, $d(t_f)$ denotes the future target--observer distance. These questions extend spatial reasoning from observation to prediction, assessing the model's ability to integrate the spatial and temporal understanding developed in the preceding levels into coherent predictions of future spatial states.

\noindent\textbf{Dataset and Benchmark.}
\label{sec:benchmark}
We construct SpatialMind-30K Dataset with 30,677 training QA pairs and spatial metadata, and SpatialMind-2K Benchmark with 2,000 questions. Figure~\ref{fig:data} shows the benchmark's task distribution. We review benchmark questions and answers through a human-in-the-loop process, using geometric and temporal computation records to correct errors. Across three progressive levels and eight task categories, the dataset and benchmark support systematic training and evaluation of predictive spatial reasoning at real-world scale. Please refer to Appendix~\ref{app:audited_data_engine} for further details on the data engine and Appendix~\ref{app:audited_tasks} for the mathematical introduction of all tasks.

\section{SpatialMind Model}
\label{sec:model}

Building on the data set and the benchmark constructed in Sect.~\ref{sec:data_engine}, we introduce SpatialMind, a VLM with an absolute metric depth adapter for real-world scale reasoning and a progressive state chain that guides the model to establish current spatial states, characterize observed dynamics, and predict future spatial conditions, grounding the final answer in a sequence of explicit state estimates.

\begin{figure}[t]
    \centering
    \includegraphics[width=1.0\linewidth]{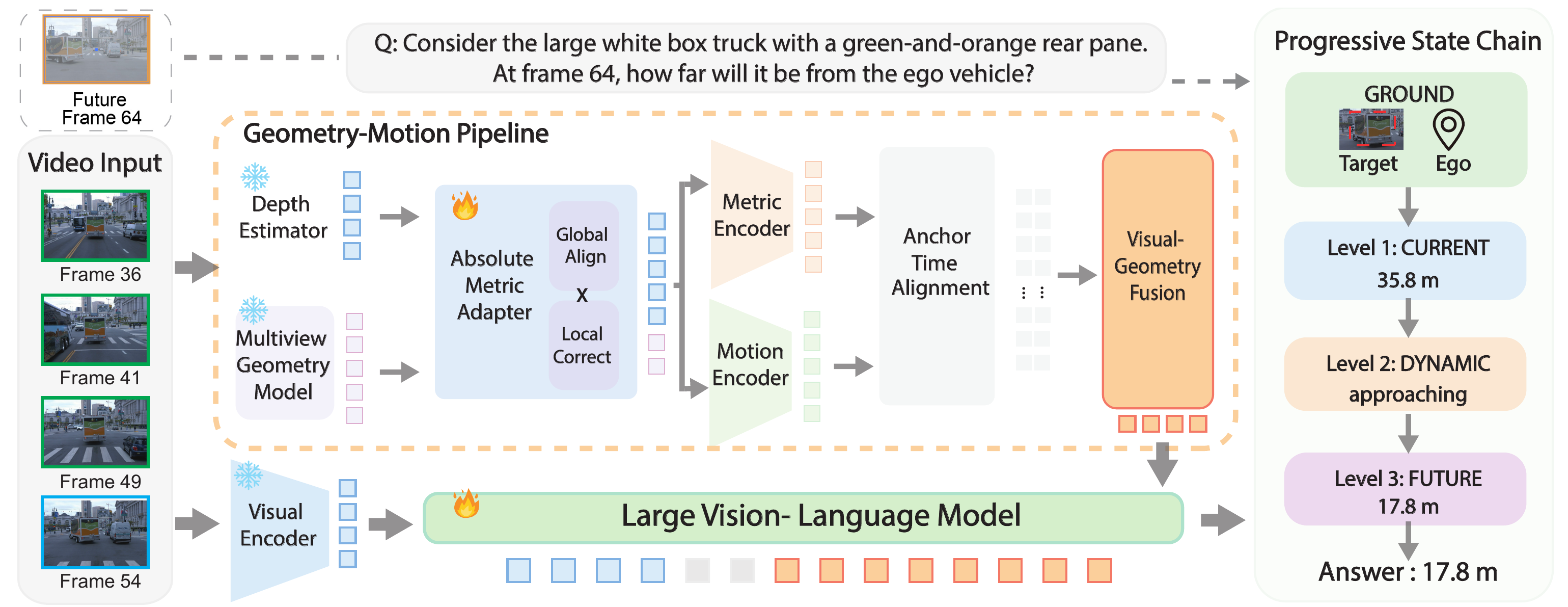}
    \vspace{-2em}
    \caption{
    \textbf{Model Architecture.} Given an observed RGB video and a spatial question, SpatialMind refines metric depth with multiview geometry through global and local corrections. We separately encode metric geometry and camera motion, align their features temporally, and integrate them into visual tokens through cross-attention. The VLM then generates a progressive state chain that grounds the target, establishes its current spatial state, and extends to observed dynamics and future prediction according to the question level before producing the final answer.
    }
    \vspace{-1em}

    \label{fig:model}
\end{figure}

\subsection{Metric Depth Adaptation and Spatiotemporal Fusion}
\label{sec:metric_adapter}
\label{sec:metric_motion_features}

To reason about distances and changes in motion, a VLM needs to estimate scene geometry at real-world scale from visual observations. However, scale errors in pretrained geometric predictions can affect distance estimates and subsequent reasoning. We introduce an absolute metric depth adapter to refine metric depth, followed by spatiotemporal fusion to integrate geometry over time.

\noindent\textbf{Absolute Metric Depth Adapter.}
As shown in Figure~\ref{fig:model}, the adapter combines initial metric depth $D_0$ from a pretrained depth estimator with multiframe features, depth maps, and confidence maps from a pretrained multiview geometry model. It predicts a global scale correction $\delta_{\mathrm g}$ for each frame and a local depth correction map $\Delta_{\mathrm l}$ for individual pixels. Applying both corrections to $D_0$ in logarithmic depth space yields the refined metric depth prediction $D_{\mathrm m}$:
\begin{equation}
    D_{\mathrm m}=D_0\odot\exp\left(\gamma_{\mathrm g}\tanh\delta_{\mathrm g}+\mathcal U\left(\gamma_{\mathrm l}\tanh\Delta_{\mathrm l}\right)\right),
    \label{eq:metric_adapter}
\end{equation}
where $\mathcal U$ upsamples the local correction to the depth resolution, and $\odot$ denotes multiplication at corresponding pixels. The coefficients $\gamma_{\mathrm g}$ and $\gamma_{\mathrm l}$ bound the corrections to limit deviations from $D_0$. We train the adapter with dense metric depth supervision while keeping both pretrained models frozen, then freeze it during VLM training. Further details are provided in Appendix~\ref{app:model_implementation}.

\noindent\textbf{Spatiotemporal Fusion.}
\label{sec:temporal_geometry_motion}
The Metric Encoder processes geometry derived from $D_{\mathrm m}$, while the Motion Encoder processes relative camera changes from the multiview model. To align these cues with video content, we extract them from the same sampled RGB frames and use their frame times for Anchor Time Alignment, as shown in Figure~\ref{fig:model}. Cross attention then fuses the encoded geometry and motion into visual tokens, providing spatial and motion cues for the progressive state chain.

\subsection{Progressive State Chain}
\label{sec:state_chain_learning}

To predict future spatial states, a VLM needs to relate current states to observed motion. However, final answer supervision does not explicitly constrain these intermediate estimates. We introduce a progressive state chain to supervise these estimates according to each question's temporal scope.

As shown in Figure~\ref{fig:model}, we train the same VLM to generate chains of different lengths for the three task levels. Level~1 includes \texttt{GROUND} to identify the target and spatial reference, and \texttt{CURRENT} to describe spatial and motion states. Level~2 adds \texttt{DYNAMICS} for observed changes, while Level~3 adds \texttt{FUTURE} for future predictions. Each chain ends with \texttt{ANSWER} to provide the final response. The model generates each state section conditioned on the video, question, and preceding sections.

We derive supervision for each state section from spatial metadata in Sec.~\ref{sec:data_engine}. Current states and observed dynamics use annotations from the observed interval. For future questions, annotations from the withheld interval provide future state targets, while the video input is the observed prefix.

\noindent\textbf{Training and Inference.}
We train the VLM with LoRA on state chain and direct answer examples from the same QA pairs. Let $X$ denote the fused video representation, $q$ the question, $a$ the answer, and $S_{1:m}$ the state sections before \texttt{ANSWER}, with $m=2,3,4$ for Levels~1, 2, and 3. The losses are
\begin{equation}
    \mathcal L_{\mathrm S}
    =\frac{1}{m}\sum_{k=1}^{m}\ell_\theta(S_k\mid X,q,S_{<k}),
    \qquad
    \mathcal L_{\mathrm A}
    =\ell_\theta(a\mid X,q,S_{1:m}).
    \label{eq:state_answer_losses}
\end{equation}
Here, $\ell_\theta$ averages token cross entropy over each target sequence, with $\theta$ denoting VLM parameters and $S_{<k}$ the preceding state sections. Averaging over sections gives them equal weight regardless of length. The direct answer loss $\mathcal L_{\mathrm D}$ uses the same averaging without a state chain. The total loss is
\begin{equation}
    \mathcal L=\rho\mathcal L_{\mathrm D}+(1-\rho)\left[(1-\lambda)\mathcal L_{\mathrm A}+\lambda\mathcal L_{\mathrm S}\right].
    \label{eq:training_objective}
\end{equation}
Here, $\rho$ balances direct answer and chain supervision, while $\lambda$ balances state and answer losses within chain examples. At inference, SpatialMind receives only the observed RGB video and question, then generates the state sections for the question's task level, followed by the final answer.

\section{Experiments}

\begin{table}[!t]
    \centering
    \caption{\textbf{Results on SpatialMind-Benchmark (\%).}
    Tasks are grouped by reasoning level. Among evaluations within each group,
    \textbf{bold} marks the best result and \underline{underline} marks the second best.}
    \label{tab:spatialmind_main}

    \begingroup
    \fontsize{8}{9.2}\selectfont
    \setlength{\tabcolsep}{1.8pt}
    \renewcommand{\arraystretch}{1.06}

    \begin{tabularx}{\linewidth}{l*{9}{>{\centering\arraybackslash}X}}
        \toprule
        \rowcolor{SMHeader}
        & &
        \multicolumn{2}{c}{\makebox[0pt][c]{\textbf{L1: Current state}}}
        & \multicolumn{3}{c}{\makebox[0pt][c]{\textbf{L2: Observed dynamics}}}
        & \multicolumn{3}{c}{\makebox[0pt][c]{\textbf{L3: Future prediction}}} \\
        \cmidrule(lr){3-4}
        \cmidrule(lr){5-7}
        \cmidrule(lr){8-10}
        Model & Overall & Motion & Spatial & Ego & Object & Relative
        & Direc. & Dist. & Relat. \\
        \midrule

        \rowcolor{SMGroup}
        \multicolumn{10}{l}{\SMgroup{Proprietary Models (API)}}\\

        GPT-5.6 Luna
        & 22.60 & 20.28 & 14.34 & 12.59 & 25.87 & 44.76 & 27.27 & 14.04 & 25.61 \\

        Claude Sonnet 5
        & 27.80 & 16.43 & \textbf{23.08} & 9.79 & 26.57 & 49.30 & \underline{30.07} & 23.51 & \underline{44.21} \\

        Gemini 3.8 Flash
        & \underline{30.19} & \underline{21.62} & \underline{18.66} & \underline{19.39} & \underline{49.65} & \underline{52.85} & 20.28 & \textbf{41.34} & 22.91 \\

        GPT-6-Astra (medium)
        & \textbf{39.00} & \textbf{38.46} & 13.99 & \textbf{19.58} & \textbf{51.75} & \textbf{73.78} & \textbf{39.86} & \underline{29.12} & \textbf{52.28} \\

        \midrule
        \rowcolor{SMGroup}
        \multicolumn{10}{l}{\SMgroup{General-purpose VLMs}}\\

        Qwen2.5-VL-3B~\citep{bai2025qwen25vltechnicalreport}
        & 14.40 & \underline{5.95} & 5.25 & \textbf{10.85} & 2.80 & 32.20 & 28.70 & 17.85 & 13.00 \\

        Qwen3.5-9B~\citep{qwen35blog}
        & 10.30 & \textbf{10.85} & 9.45 & 5.60 & 13.30 & 9.45 & 4.20 & \underline{20.35} & 8.05 \\

        LLaVA-NeXT-7B~\citep{zhang2024llavanextvideo}
        & 14.40 & 5.25 & 6.65 & 5.95 & 7.00 & 30.10 & \underline{31.50} & 15.05 & 18.60 \\

        Qwen2.5-VL-7B~\citep{bai2025qwen25vltechnicalreport}
        & 17.60 & 3.85 & \textbf{17.15} & 2.45 & \underline{30.80} & 33.60 & 25.90 & 10.15 & 27.70 \\

        Cosmos3-8B~\citep{agarwal2026cosmos}
        & \underline{18.95} & 5.60 & 2.40 & \underline{9.80} & \textbf{32.90} & \underline{35.30} & \textbf{32.20} & 3.90 & \textbf{43.20} \\

        Qwen3-VL-8B~\citep{bai2025qwen3}
        & \textbf{20.20} & 3.15 & \underline{13.30} & 7.35 & 23.10 & \textbf{38.50} & 24.50 & \textbf{21.00} & \underline{34.70} \\

        \midrule
        \rowcolor{SMGroup}
        \multicolumn{10}{l}{\SMgroup{Spatially specialized VLMs}}\\

        SR-3D-8B~\citep{cheng20263d}
        & 9.60 & 2.45 & 6.30 & \underline{12.60} & 2.10 & 24.85 & 14.70 & \underline{8.75} & 3.50 \\

        SpatialRGPT-8B~\citep{cheng2024spatialrgpt}
        & 9.70 & 6.65 & 5.25 & 5.25 & 2.80 & 17.85 & \textbf{34.30} & 3.85 & 10.20 \\

        SpatialBot-3B~\citep{cai2025spatialbot}
        & 11.80 & 2.80 & 7.70 & 1.05 & \textbf{34.30} & 23.45 & 28.70 & 2.80 & 13.00 \\

        VLM-3R-7B~\citep{fan2026vlm}
        & 14.00 & \textbf{7.90} & 1.05 & 4.20 & 7.70 & 31.50 & 29.40 & 2.45 & \textbf{32.65} \\

        SpaceR-7B~\citep{ouyang2025spacer}
        & 14.70 & 5.60 & \textbf{14.70} & 5.25 & \underline{14.70} & \underline{33.25} & 21.70 & 3.50 & 22.45 \\

        ViLaSR-7B~\citep{wu2026reinforcing}
        & 14.80 & \underline{7.70} & 9.10 & 7.00 & 7.00 & 32.20 & 16.80 & 6.00 & \underline{30.15} \\

        SR-ReaL-8B~\citep{ji2026reinforcing}
        & 14.90 & 5.95 & 11.55 & 11.90 & 8.40 & 31.50 & \underline{32.20} & 7.05 & 16.15 \\

        4D-RGPT-8B~\citep{yang20264d}
        & \underline{15.40} & 3.50 & \underline{11.90} & \textbf{12.95} & 7.00 & \textbf{33.95} & 27.30 & \textbf{12.25} & 16.15 \\

        SpatialReasoner-7B~\citep{ma2026spatialreasoner}
        & \textbf{15.80} & 7.20 & 7.70 & 5.60 & 12.60 & \underline{33.25} & 30.10 & 8.40 & 27.00 \\

        \midrule
        \rowcolor{SMGroup}
        \multicolumn{10}{l}{\SMgroup{SFT on SpatialMind training data}}\\

        Qwen3-VL-8B + SFT
        & \underline{45.70} & \underline{36.02} & \underline{57.34} & \underline{23.77} & \underline{37.06} & \underline{56.99} & \underline{57.34} & \underline{40.64} & \underline{57.89} \\

        \rowcolor{SMOurs}
        \textbf{Ours}
        & \textbf{52.65} & \textbf{38.32} & \textbf{65.46} & \textbf{30.15} & \textbf{52.34} & \textbf{68.42} & \textbf{65.13} & \textbf{42.89} & \textbf{64.48} \\

        \bottomrule
    \end{tabularx}
    \endgroup
\end{table}

\noindent\textbf{Baselines and Setup.}
To evaluate predictive spatial reasoning, we compare SpatialMind with proprietary VLMs, including GPT, Claude, and Gemini models; general-purpose models such as Qwen3-VL~\citep{bai2025qwen3} and Cosmos~3~\citep{agarwal2026cosmos}; and spatially specialized models such as SpatialRGPT~\citep{cheng2024spatialrgpt} and VLM-3R~\citep{fan2026vlm}. We also fine-tune Qwen3-VL-8B on SpatialMind-30K as an SFT baseline using the same backbone and training data. For future-prediction tasks, all models receive only the observed video prefix and question.

\noindent\textbf{Evaluation Metrics.}
SpatialMind-2K Benchmark contains three multiple-choice categories and five numerical fill-in-the-blank categories. We evaluate multiple-choice questions on relative motion, ego future direction, and target future relation using accuracy (ACC). The numerical categories cover spatial state, motion state, ego-motion distance, target-motion distance, and target future distance. We evaluate numerical answers, such as absolute distances, using mean relative accuracy (MRA), following prior spatial reasoning evaluations~\citep{yang2025thinking,fan2026vlm}.

\subsection{Comparison on SpatialMind-Benchmark}
\label{sec:benchmark_comparison}

We compare SpatialMind with models in four categories: proprietary models, general-purpose open-source VLMs, spatially specialized VLMs, and a baseline fine-tuned on SpatialMind-Dataset. SpatialMind-Benchmark evaluates these models across three progressive levels, from understanding current spatial states to reasoning about observed dynamics and predicting future states.

\noindent\textbf{Proprietary Models.}
We first evaluate proprietary models through their APIs, including GPT-5.6 Luna, Claude Sonnet 5, Gemini 3.8 Flash, and GPT-6-Astra.
As shown in Table~\ref{tab:spatialmind_main}, these models achieve higher overall accuracy than the evaluated general-purpose and spatially specialized VLMs.
All four achieve their highest task accuracy on observed relative motion, while current-state distance and ego-motion distance estimation remain less accurate.
These results suggest that stronger overall performance can coexist with difficulties in estimating distances at real-world scale.

\noindent\textbf{General-Purpose VLMs.}
We evaluate representative general-purpose VLMs, including Qwen3-VL~\citep{bai2025qwen3}, LLaVA~\citep{zhang2024llavanextvideo}, and Cosmos~3~\citep{agarwal2026cosmos}, which support broad vision-language understanding. As shown in Table~\ref{tab:spatialmind_main}, these models tend to achieve higher accuracy on categorical questions involving directions or spatial relations than on tasks requiring direct estimation of metric quantities such as distance, displacement, and speed. Metric estimation remains challenging in Level~1, when establishing the current spatial state, and Level~2, particularly when measuring ego-motion distance. We further observe that some models retain relatively stronger performance on categorical future-prediction questions despite less accurate metric estimation at the preceding levels. The results suggest that stronger performance on categorical future questions does not necessarily imply accurate metric understanding of the observed scene.

\noindent\textbf{Spatially Specialized VLMs.}
We further evaluate VLMs specialized for spatial reasoning, many of which incorporate depth or 3D geometric priors to enhance spatial perception. As shown in Table~\ref{tab:spatialmind_main}, these models remain limited on metric tasks, particularly Level~1, ego-motion distance in Level~2, and future-distance prediction in Level~3. SpatialMind-Benchmark spans outdoor driving scenes and everyday egocentric environments, requiring models to estimate distances and motion across different spatial scales. The results suggest that incorporating geometric priors alone does not ensure accurate metric reasoning across the diverse scenes covered by our benchmark.

\noindent\textbf{SpatialMind.}
We first fine-tune Qwen3-VL-8B on SpatialMind-30K dataset to establish an SFT baseline. Compared with zero-shot Qwen, this baseline improves performance across all eight tasks, with substantial gains in metric estimation and motion reasoning, suggesting the benefit of supervision grounded in linguistic entities, metric geometry, and temporal computations. We then train our SpatialMind model on the same dataset and observe further improvements across all tasks, increasing overall accuracy from $45.70\%$ to $52.65\%$. These additional gains support the combined benefit of incorporating metric geometry and structuring inference through the progressive state chain, which establishes the current spatial state, summarizes observed dynamics, and predicts future states.

\subsection{Generalization to Spatial Benchmarks}

To examine the generalization of SpatialMind beyond our benchmark, we conduct
zero-shot evaluation on three complementary spatial reasoning benchmarks.
VSI-Bench~\citep{yang2025thinking}  evaluates video-based spatial understanding, with emphasis on metric
measurement and spatiotemporal reasoning; OSI-Bench~\citep{wu2026indoor} focuses on open-world
spatial inference in both static and dynamic metric settings; and VLM4D~\citep{zhou2025vlm4d}
evaluates 4D spatial-temporal understanding over real and synthetic videos.
Together, these benchmarks cover spatial measurement, dynamic reasoning, and
motion understanding across diverse scenes and evaluation protocols.

\noindent\textbf{VSI-Bench.}
As shown in Table~\ref{tab:external_results}, SpatialMind achieves
the highest measurement and spatiotemporal scores among the compared
models, reaching 52.1 and 52.4. These exceed the strongest competing
scores by 4.4 and 3.8 points, respectively, while the overall score
of 46.4 remains competitive. The results support the transfer of
metric estimation and spatiotemporal reasoning to indoor scenes.

\noindent\textbf{OSI-Bench.}
SpatialMind achieves the highest overall score of 35.0 among the
compared models, surpassing Qwen3-VL-32B by 2.8 points. It leads in
static metric reasoning with 30.2 and obtains a competitive dynamic
metric score of 33.4, compared with 34.9 for Qwen3-VL-32B. These
results support the transfer of static and dynamic metric reasoning
to open-world environments.

\noindent\textbf{VLM4D.}
SpatialMind achieves an overall score of 53.4, exceeding 4D-RGPT by
1.0 points, and obtains the highest scores among the compared models
on both real and synthetic videos, reaching 53.5 and 52.9, respectively.
The gains across both splits provide further evidence that SpatialMind
generalizes to motion and perspective understanding across different
visual environments.

\subsection{Ablation Studies}
We examine the metric depth adaptation pipeline and progressive state chain introduced in Sec.~\ref{sec:model} to assess their contributions to SpatialMind.
We compare geometric inputs and reasoning formats across the three reasoning levels to evaluate their effects on predictive spatial reasoning.

\noindent\textbf{Geometric Inputs and Metric Adaptation.}
We evaluate the metric depth adaptation pipeline in Sec.~\ref{sec:metric_adapter} by comparing different geometric inputs and their effects on spatial reasoning accuracy.
The \emph{RGB-only} variant uses visual features without geometric conditioning.
The \emph{DA3} variant replaces our metric depth adaptation pipeline with depth predicted directly by DA3~\citep{lin2025depth}.
The \emph{VGGT-$\Omega$} variant independently replaces the same pipeline with depth predicted by VGGT-$\Omega$~\citep{wang2026vggt}.
The full model uses the adapted metric depth produced by our pipeline, and all variants retain the progressive state chain.
As shown in Table~\ref{tab:geometry_ablation}, both pretrained-depth alternatives provide modest gains over RGB-only input, while the full model achieves the highest accuracy overall and across all three reasoning levels.
The largest improvement over RGB-only input occurs in Level~2 observed-dynamics reasoning, which requires estimating spatial changes across frames.
These results support the effectiveness of our complete metric depth adaptation pipeline.

\noindent\textbf{Progressive State-Chain Reasoning.}
We evaluate the progressive state chain by comparing different output formats and their effects on spatial reasoning accuracy.
The \emph{direct-answer} variant trains the model to output only the final answer, without an intermediate state chain.
For \emph{free-form CoT}, we prompt the original Qwen3-VL model to follow a \texttt{detect--think--answer} format without task-specific fine-tuning.
The \emph{w/o grounding} variant removes the initial \texttt{GROUND} step, which interprets the question by identifying the target, spatial reference, and quantity to estimate.
The direct-answer, w/o grounding, and full-chain variants retain the same geometric pipeline.
As shown in Table~\ref{tab:state_chain_ablation}, the full chain improves accuracy over direct-answer prediction across all three reasoning levels and also exceeds the prompt-based CoT reference.
Removing \texttt{GROUND} lowers overall accuracy and performance on Levels~1 and~2, while yielding a slightly higher Level~3 score.
These results support the benefit of state-chain supervision for predictive spatial reasoning.

\begin{table}[!t]
    \vspace{-1.0em}
    \centering
    \caption{\textbf{Zero-shot evaluation on external benchmarks.}
    VSI-Bench reports measurement (Meas.) and spatiotemporal (S-T) reasoning;
    OSI-Bench reports static and dynamic reasoning;
    VLM4D reports the real and synthetic splits. Our method achieves competitive results across three benchmarks.}
    \label{tab:external_results}

    \begingroup
    \fontsize{7.2}{8.3}\selectfont
    \setlength{\tabcolsep}{1.5pt}
    \renewcommand{\arraystretch}{1.10}

    \begin{minipage}[t]{0.36\linewidth}
        \centering

        \begin{tabularx}{\linewidth}{lYYY}
            \toprule
            \rowcolor{SMHeader}
            \multicolumn{4}{c}{\textbf{VSI-Bench}} \\
            \midrule
            Model & Avg. & Meas. & S-T \\
            \midrule

            GPT-4o
            & 34.0
            & 29.1
            & 28.5 \\

            Gemini-1.5-Pro
            & 45.4
            & 46.2
            & 34.6 \\

            InternVL2-8B
            & 37.5
            & 40.7
            & 46.4 \\

            LLaVA-NeXT-Video-72B
            & 40.9
            & 38.5
            & 48.6 \\

            LLaVA-OneVision-72B
            & 40.2
            & 39.7
            & 44.6 \\

            Spatial-MLLM-4B
            & \textbf{48.4}
            & 47.7
            & 46.3 \\

            \rowcolor{SMOurs}
            \textbf{Ours}
            & 46.4
            & \textbf{52.1}
            & \textbf{52.4} \\

            \bottomrule
        \end{tabularx}

    \end{minipage}
    \hfill
    \begin{minipage}[t]{0.32\linewidth}
        \centering

        \begin{tabularx}{\linewidth}{lYYY}
            \toprule
            \rowcolor{SMHeader}
            \multicolumn{4}{c}{\textbf{OSI-Bench}} \\
            \midrule
            Model & Avg. & Static & Dyn. \\
            \midrule

            GPT-5
            & 29.7
            & 25.7
            & 25.0 \\

            Claude-3.7-Sonnet
            & 26.5
            & 28.4
            & 13.4 \\

            InternVL3.5-8B
            & 28.5
            & 22.6
            & 24.0 \\

            Ovis2-16B
            & 28.1
            & 22.6
            & 24.4 \\

            Qwen3-VL-8B
            & 31.2
            & 23.1
            & 31.1 \\

            Qwen3-VL-32B
            & 32.2
            & 22.3
            & \textbf{34.9} \\

            \rowcolor{SMOurs}
            \textbf{Ours}
            & \textbf{35.0}
            & \textbf{30.2}
            & 33.4 \\

            \bottomrule
        \end{tabularx}

    \end{minipage}
    \hfill
    \begin{minipage}[t]{0.30\linewidth}
        \centering

        \begin{tabularx}{\linewidth}{l
            >{\hsize=1.15\hsize\linewidth=\hsize}Y
            >{\hsize=0.85\hsize\linewidth=\hsize}Y
            Y}
            \toprule
            \rowcolor{SMHeader}
            \multicolumn{4}{c}{\textbf{VLM4D}} \\
            \midrule
            Model & Overall & Real & Synth. \\
            \midrule

            Qwen3-VL-8B
            & 52.2
            & 52.1
            & 52.4 \\

            Claude-Sonnet-4
            & 51.3
            & 52.2
            & 48.3 \\

            InternVL2.5-8B
            & 42.0
            & 42.4
            & 40.9 \\

            VG-LLM-7B
            & 46.5
            & 49.5
            & 37.3 \\

            VLM-3R-7B
            & 33.9
            & 36.9
            & 24.7 \\

            4D-RGPT-8B
            & 52.4
            & 53.1
            & 50.6 \\

            \rowcolor{SMOurs}
            \textbf{Ours}
            & \textbf{53.4}
            & \textbf{53.5}
            & \textbf{52.9} \\

            \bottomrule
        \end{tabularx}

    \end{minipage}

    \endgroup
\end{table}

\begin{table}[!t]
    \vspace{-1.1em}
    \centering
    \begin{minipage}[t]{0.49\linewidth}
        \centering
        \caption{\textbf{Geometric ablation}
        with the progressive state chain used in all variants.}
        \label{tab:geometry_ablation}
        \begingroup
        \fontsize{8}{9.2}\selectfont
        \setlength{\tabcolsep}{2.2pt}
        \renewcommand{\arraystretch}{1.08}

        \begin{tabularx}{\linewidth}{lYYYY}
            \toprule
            \rowcolor{SMHeader}
            Variant & Overall & Level~1 & Level~2 & Level~3 \\
            \midrule

            RGB only
            & 50.42
            & 48.96
            & 46.31
            & 54.42 \\

            + DA3
            & 51.30
            & 49.48
            & 48.13
            & 55.59 \\

            + $\Omega$
            & 51.20
            & 50.17
            & 46.62
            & 55.71 \\

            \rowcolor{SMOurs}
            Full model
            & \SMbest{52.65}
            & \SMbest{51.89}
            & \SMbest{50.30}
            & \SMbest{57.47} \\

            \bottomrule
        \end{tabularx}

        \endgroup
    \end{minipage}
    \hfill
    \begin{minipage}[t]{0.49\linewidth}
        \centering
        \caption{\textbf{Progressive state-chain ablation}
        with absolute metric adapter retained in all variants.}
        \label{tab:state_chain_ablation}
        \begingroup
        \fontsize{8}{9.2}\selectfont
        \setlength{\tabcolsep}{2.2pt}
        \renewcommand{\arraystretch}{1.08}

        \begin{tabularx}{\linewidth}{lYYYY}
            \toprule
            \rowcolor{SMHeader}
            Variant & Overall & Level~1 & Level~2 & Level~3 \\
            \midrule

            Direct answer
            & 49.95
            & 48.43
            & 48.25
            & 55.22 \\

            Free-form CoT
            & 34.15
            & 33.05
            & 35.57
            & 33.73 \\

            w/o grounding
            & 51.55
            & 49.48
            & 46.74
            & \SMbest{57.58} \\

            \rowcolor{SMOurs}
            Full state chain
            & \SMbest{52.65}
            & \SMbest{51.89}
            & \SMbest{50.30}
            & 57.47 \\

            \bottomrule
        \end{tabularx}

        \endgroup
    \end{minipage}
    \vspace{-1.0em}
\end{table}

\section{Conclusion}

We introduced SpatialMind, a spatial reasoning VLM that extends scene
understanding to physically grounded future prediction. We formulate spatial
reasoning as a progressive process that establishes current spatial states,
captures observed dynamics, and uses both to anticipate future states beyond
the observed interval. To support this progression, we develop a scalable
data engine and construct SpatialMind-30K Dataset and SpatialMind-2K Benchmark
with metric geometric and temporal supervision. SpatialMind combines an
absolute metric depth adapter with a progressive state chain to ground
reasoning at real-world scale and structure inference from observation to
prediction. Experiments demonstrate consistent gains across all three
reasoning levels and competitive zero-shot performance on external benchmarks,
while ablations support the contributions of both components. These findings
highlight the value of metric grounding and observed-dynamics understanding
for enabling VLMs to anticipate future spatial changes in dynamic environments.

\subsection*{AI Use Statement}
\vspace{-0.5em}
Generative AI assistants were used only to polish the manuscript for clarity, grammar, concision, and presentation. Proprietary generative AI models were evaluated solely as baselines on our benchmark and were not used as components of our method. All AI-assisted edits were reviewed by the authors, who take responsibility for the final content of the paper.

\subsection*{Ethics Statement}
\vspace{-0.5em}
This work uses the Waymo Open Dataset and Ego-Exo4D under their applicable
access and usage terms. Our dataset and benchmark may inherit biases
and privacy limitations from the source data. SpatialMind is intended for research purposes only, supporting the study
of spatial reasoning and future prediction in vision-language models.

\subsection*{Reproducibility Statement}
\vspace{-0.5em}
We provide detailed information required to reproduce our method and
experiments in the appendix, including the data construction pipeline,
task definitions and geometric computations, model implementation and
training details, dataset composition, and evaluation protocol.
Additional worked examples illustrate the construction of reference
answers and progressive state-chain supervision.
We will release the
code, dataset, and evaluation tools to facilitate reproduction and
further research.

\bibliography{spatialmind}
\bibliographystyle{spatialmind}

\clearpage
\appendix
\begingroup
\microtypesetup{activate=true}
\addtolength{\textwidth}{0.25in}
\addtolength{\oddsidemargin}{-0.125in}
\addtolength{\evensidemargin}{-0.125in}
\setlength{\linewidth}{\textwidth}
\setlength{\columnwidth}{\textwidth}
\setlength{\hsize}{\textwidth}
\setlength{\headwidth}{\textwidth}
\tolerance=2000
\renewcommand{\baselinestretch}{0.96}
\setlength{\parskip}{2pt plus 0.5pt minus 0.5pt}
\setlength{\medskipamount}{1pt plus 0.5pt minus 0.5pt}
\setlength{\smallskipamount}{2pt}
\setlength{\abovedisplayskip}{3pt plus 1pt minus 1pt}
\setlength{\belowdisplayskip}{3pt plus 1pt minus 1pt}
\setlength{\abovedisplayshortskip}{0pt}
\setlength{\belowdisplayshortskip}{2pt plus 1pt minus 1pt}
\setlength{\intextsep}{8pt plus 1pt minus 1pt}
\setlength{\textfloatsep}{10pt plus 1pt minus 1pt}
\setlength{\floatsep}{8pt plus 1pt minus 1pt}
\setlength{\abovecaptionskip}{6pt}

\let\SMAppendixNormalSize\normalsize
\renewcommand{\normalsize}{
    \SMAppendixNormalSize
    \setlength{\abovedisplayskip}{3pt plus 1pt minus 1pt}
    \setlength{\belowdisplayskip}{3pt plus 1pt minus 1pt}
    \setlength{\abovedisplayshortskip}{0pt}
    \setlength{\belowdisplayshortskip}{2pt plus 1pt minus 1pt}
}
\normalsize

\makeatletter
\def\section{\@startsection{section}{1}{\z@}{-1.5ex plus -.3ex minus -.2ex}{1ex plus .2ex}{\large\sc\raggedright}}
\def\subsection{\@startsection{subsection}{2}{\z@}{-1.4ex plus -.3ex minus -.2ex}{.6ex plus .2ex}{\normalsize\sc\raggedright}}
\def\paragraph{\@startsection{paragraph}{4}{\z@}{.8ex plus .2ex minus .2ex}{-1em}{\normalsize\bf}}
\makeatother

\begin{center}
    {\LARGE\bfseries Supplementary Material}
\end{center}

\noindent
This supplementary material provides further details on the SpatialMind
data engine, geometric computations for the eight task families, and the
metric depth adapter and temporal fusion. We also document the dataset
composition and evaluation metrics, analyze the persistence of current
states in future-prediction questions, and discuss related spatiotemporal
tasks and limitations. Finally, worked examples and visualizations
illustrate how video observations and geometric annotations are used to
construct reference answers and progressive state-chain supervision.
\looseness=-1

\medskip
\noindent\textbf{Contents}
\par\smallskip

\begingroup
\renewcommand{\arraystretch}{1.0}
\noindent
\begin{tabular*}{\linewidth}{
    @{}l@{\hspace{0.8em}}l@{\extracolsep{\fill}}r@{}
    }

    \ref{app:audited_data_engine} &
    Data Engine Details &
    \pageref{app:audited_data_engine} \\

    \ref{app:audited_tasks} &
    Task Definitions and Geometric Computations &
    \pageref{app:audited_tasks} \\

    \ref{app:model_implementation} &
    Metric Geometry and Temporal Fusion &
    \pageref{app:model_implementation} \\

    \ref{app:integrity} &
    Dataset Composition and Evaluation &
    \pageref{app:integrity} \\

    \ref{app:additional_analyses} &
    Additional Analysis &
    \pageref{app:additional_analyses} \\

    \ref{app:related_scope} &
    Relationship to Other Spatiotemporal Tasks &
    \pageref{app:related_scope} \\

    \ref{app:limitations} &
    Limitations and Future Work &
    \pageref{app:limitations} \\

    \ref{app:visualizations} &
    Visualizations &
    \pageref{app:visualizations} \\

\end{tabular*}
\endgroup

\medskip

\section{Data Engine Details}
\label{app:audited_data_engine}

\subsection{Source Geometry and Correspondences}
\label{app:source_geometry}
Waymo~\citep{sun2020scalability} supplies sparse LiDAR measurements, calibrated cameras, observer poses, and tracked 3D boxes. For Ego-Exo4D~\citep{grauman2024ego}, Aria semidense landmarks and their image observations provide sparse metric references, together with the device trajectory and camera calibration. The references are aligned with the RGB images by composing the relevant sensor transformations. Source track identifiers establish correspondences when available; otherwise, SAM3~\citep{carion2026sam} video segmentation propagates instance identities. Maintaining such identities over time is also studied using continuous
spatiotemporal instance fields~\citep{wu2026consistent}. Following Section~\ref{sec:metric_annotation}, calibrated depth within each selected target mask is back-projected using camera intrinsics and poses into the shared world frame $W$. The resulting per-frame 3D localizations are associated over time to obtain target trajectories. The observer's position is the camera centre, and its heading is given by the camera orientation.
\looseness=-1

\subsection{Dense-depth Calibration}
\label{app:engine_depth}
DA3-Large~\citep{lin2025depth}, the Large variant of Depth Anything 3 (DA3), predicts dense depth, and sparse metric references determine its scale and offset. At frame $t$, a reference pixel $\mathbf x_i$ pairs the predicted depth $\widehat D_t(\mathbf x_i)$ with the metric measurement $z_{t,i}$. We fit the affine relation $z_{t,i}\simeq a_t\widehat D_t(\mathbf x_i)+b_t$ using valid, positive pairs, reducing the influence of inconsistent references through robust weighting. The fitted coefficients are stabilized across frames. Denoting the resulting coefficients by $\bar a_t$ and $\bar b_t$, the calibrated depth is
\looseness=-1
\begin{equation}
    D_t(\mathbf x)=\max\{0,\bar a_t\widehat D_t(\mathbf x)+\bar b_t\}.
    \label{eq:app:engine_depth_output}
\end{equation}
Here $D_t$ and the offset $\bar b_t$ are expressed in metres. Valid calibrated depths provide the visible-surface geometry used with instance masks, while sparse references support calibration checks. This procedure supplies the metric geometry described in Section~\ref{sec:metric_annotation}.
\looseness=-1

\subsection{Referring-expression Construction}
\label{app:engine_referring}
Following Section~\ref{sec:linguistic_grounding}, we select sufficiently visible targets and use Qwen3-VL-32B~\citep{bai2025qwen3} to describe their category, appearance, and distinguishing attributes. Image location may disambiguate visually similar instances. Explicit metric distances, motion states, and future relations are excluded from the descriptions. Reverse localization then uses the description and an unmarked frame to identify the target and check alternative matches. Incorrect or ambiguous descriptions are revised or discarded. Accepted descriptions remain linked to persistent entity identifiers, connecting the question's referent to its geometry throughout the video.
\looseness=-1

\section{Task Definitions and Geometric Computations}
\label{app:audited_tasks}
The eight task families follow Section~\ref{sec:qa_generation}. We retain its notation: $\mathbf p_u^W(t)$ is the position of entity $u$ in metres in the shared world frame $W$, with $u=\mathrm{ego}$ for the observer and $u=\mathrm{tag}$ for the target. The variables $t,t_s,t_o,t_f$ denote integer frame indices, not elapsed seconds. A question may specify a subinterval $[t_1,t_2]\subseteq[t_s,t_o]$ of the observed video. We write $\tau_t$ for the corresponding timestamp in seconds and require increasing timestamps over a queried motion interval. Future queries satisfy $t_f>t_o$. Turning-angle answers are reported in degrees; angular calculations below state explicitly when radians are used.
\looseness=-1

\subsection{Level 1: Current-state Understanding}
\label{app:audited_tasks_l1}
\noindent\textbf{Spatial state.}
This task asks about metric separation at an observed frame. Target--observer distance is $d(t)$ in \Eqref{eq:current_spatial_state}. For two queried entities $u$ and $v$, the corresponding point-to-point distance is
\looseness=-1
\begin{equation}
    d_{u,v}(t)=\|\mathbf p_u^W(t)-\mathbf p_v^W(t)\|_2,
    \qquad t_s\leq t\leq t_o.
    \label{eq:app:spatial_distance}
\end{equation}
Both positions refer to the same frame and world coordinate system, and the distance is measured in metres. For target--observer queries, $\mathbf p_{\mathrm{ego}}^W(t)$ is the camera centre defined in Section~\ref{sec:metric_annotation}. Applying the same rigid coordinate transformation to both positions preserves this distance. Figure~\ref{fig:app:l1_pan_surface_range} illustrates an explicitly requested visible-surface quantity: the median camera-to-surface range over valid target-mask samples. This scalar surface statistic is distinct from the point-to-point distance defined above.
\looseness=-1

\medskip
\noindent\textbf{Motion state.}
This task asks about speed or turning angle over a specified local interval. With $t_1<t_2$ and $\tau_{t_2}>\tau_{t_1}$, trajectory-based mean speed is
\looseness=-1
\begin{equation}
    \bar v_u(t_1,t_2)=\frac{s_u(t_1,t_2)}{\tau_{t_2}-\tau_{t_1}},
    \qquad u\in\{\mathrm{ego},\mathrm{tag}\},
    \label{eq:app:motion_speed}
\end{equation}
where $s_u$ is the accumulated path length defined in \Eqref{eq:observed_dynamics}; thus $\bar v_u$ is in metres per second. Frame counts alone do not determine elapsed time when timestamps are nonuniform. For questions defined from sampled camera speeds, the answer is the arithmetic mean of the supplied speed samples over the queried interval (Figure~\ref{fig:app:l1_ego_average_speed}); the serialized measurement convention specifies which construction applies. For view-turn questions, let $\mathbf f_t^W$ be the unit optical forward direction expressed in the same world frame at every time. The unsigned angle between the endpoint viewing directions is
\looseness=-1
\begin{equation}
    \theta=\frac{180^\circ}{\pi}\arccos\!\left(
    \operatorname{clip}_{[-1,1]}\bigl((\mathbf f_{t_1}^W)^\top\mathbf f_{t_2}^W\bigr)\right).
    \label{eq:app:view_turn}
\end{equation}
The arccosine returns radians, so the conversion gives $\theta\in[0^\circ,180^\circ]$. This endpoint viewing-angle change does not encode a left/right sign or cumulative rotation.
\looseness=-1

\subsection{Level 2: Observed-dynamics Understanding}
\label{app:audited_tasks_l2}
\noindent\textbf{Ego-motion distance.}
The answer is $s_{\mathrm{ego}}(t_1,t_2)$ from \Eqref{eq:observed_dynamics}, evaluated over the queried observation interval. It accumulates the observer's travel along the trajectory; an interval with identical endpoints can still have nonzero path length.
\looseness=-1

\medskip
\noindent\textbf{Target-motion distance.}
The answer is $s_{\mathrm{tag}}(t_1,t_2)$ from the same equation, using the same target identity throughout the interval. Measuring positions in $W$ separates target motion from image changes caused by camera motion. Both motion-distance tasks return a numerical answer in metres. The ground-route example in Figure~\ref{fig:app:l2_cyclist_route} applies the segment-length sum to its recorded ground-plane positions; this planar measurement is distinct from a general three-dimensional trajectory length.
\looseness=-1

\medskip
\noindent\textbf{Relative motion.}
This task asks whether the target becomes closer, farther, leftward, rightward, or approximately unchanged relative to the observer. The signed change in metric separation is
\looseness=-1
\begin{equation}
    \Delta d=d(t_2)-d(t_1),
    \label{eq:app:relative_distance_change}
\end{equation}
with $d(t)$ defined by \Eqref{eq:current_spatial_state}. Negative and positive values indicate decreasing and increasing separation, respectively. In the observer's optical frame, whose axes point right, down, and forward, a target at $(X_t,Y_t,Z_t)$ with $Z_t>0$ has right-positive bearing $\alpha(t)=\operatorname{atan2}(X_t,Z_t)$ in radians. For these forward-facing observations, the bearing lies in $(-\pi/2,\pi/2)$, and
\looseness=-1
\begin{equation}
    \Delta\alpha=\alpha(t_2)-\alpha(t_1)
    \label{eq:app:relative_bearing_change}
\end{equation}
is negative for leftward change and positive for rightward change. The categorical question identifies the dominant stable change, or approximately unchanged relations when both changes are small. Visibility, noise, and trend consistency checks exclude ambiguous candidates. For driving relative-motion questions, the range is normalized using every sampled frame in the queried interval. With $N=t_2-t_1+1$, define
\looseness=-1
\begin{equation}
    \begin{aligned}
        \bar d&=\frac{1}{N}\sum_{t=t_1}^{t_2}d(t),\\
        R&=\frac{|\Delta d|}{\max(\bar d,10^{-6}\,\mathrm{m})},
        \qquad L=|\Delta\alpha|,\qquad T=\sqrt{R^2+L^2}.
    \end{aligned}
    \label{eq:app:driving_relative_normalization}
\end{equation}
Here the angular difference is in radians, so both normalized terms are dimensionless. The driving rule first assigns approximately unchanged motion when $T<0.08$. Otherwise, after checking trend consistency, $R\geq1.5L$ selects the radial class and $L\geq1.5R$ selects the lateral class; the sign of the corresponding change determines farther/closer or right/left. Candidates without a dominant stable change are excluded. Figure~\ref{fig:app:level2_relative_motion} illustrates this computation.
\looseness=-1

\subsection{Level 3: Future Prediction}
\label{app:audited_tasks_l3}
Only the observed prefix is supplied as video input. Future annotations provide the reference quantities for prediction tasks.

\noindent\textbf{Ego future direction.}
This task classifies the observer's subsequent manoeuvre over a designated interval $[t_a,t_f]$, where $t_o\leq t_a<t_f$. Let $\psi_t$ be the observer's heading in radians, with positive change corresponding to a left turn. The accumulated signed turn, expressed in degrees, is
\looseness=-1
\begin{equation}
    \Theta=\frac{180^\circ}{\pi}\sum_{t=t_a}^{t_f-1}
    \operatorname{wrap}_{[-\pi,\pi)}(\psi_{t+1}-\psi_t),
    \label{eq:app:future_turn}
\end{equation}
where $\operatorname{wrap}_{[-\pi,\pi)}$ chooses the equivalent angle in $[-\pi,\pi)$. For the driving questions, terminal speed uses the final five sampled frame intervals within the future interval:
\looseness=-1
\begin{equation}
    v_{\mathrm{end}}=\frac{s_{\mathrm{ego}}(t_f-5,t_f)}{\tau_{t_f}-\tau_{t_f-5}},
    \qquad t_f-t_a\geq5.
    \label{eq:app:terminal_speed}
\end{equation}
Here the same path-length sum is applied to annotated future positions. The timestamps in the denominator are in seconds and strictly increase. Using $v_0=0.5\,\mathrm{m/s}$, the mutually exclusive label rule is
\looseness=-1
\begin{equation}
    y=\begin{cases}
        \text{come to a stop},&v_{\mathrm{end}}<v_0,\\
        \text{turn left},&v_{\mathrm{end}}\geq v_0,\ \Theta\geq12^\circ,\\
        \text{turn right},&v_{\mathrm{end}}\geq v_0,\ \Theta\leq-12^\circ,\\
        \text{go straight},&v_{\mathrm{end}}\geq v_0,\ |\Theta|<6^\circ.
    \end{cases}
    \label{eq:app:future_direction_rule}
\end{equation}
Stopping takes priority. Non-stopping candidates with $6^\circ\leq|\Theta|<12^\circ$ are excluded. The left-positive heading convention here is distinct from the right-positive bearing convention for target-relative motion.
\looseness=-1

\medskip
\noindent\textbf{Target future distance.}
The answer is $d(t_f)$ from \Eqref{eq:future_spatial_prediction}, with both the target and observer positions evaluated at the same future frame $t_f>t_o$. The model predicts this distance in metres from the observed prefix.
\looseness=-1

\medskip
\noindent\textbf{Target future relation.}
This task predicts the target's horizontal relation to the observer at $t_f$. Let $\mathbf c^W(t_f)=\mathbf p_{\mathrm{ego}}^W(t_f)$ be the future optical camera centre and let $R_{W\to C}(t_f)$ rotate world vectors into the optical camera frame, whose axes point right, down, and forward. Camera poses and calibrated sensor extrinsics determine these quantities. Project the target's future position as
\looseness=-1
\begin{equation}
    \begin{aligned}
        (X_f,Y_f,Z_f)^\top
        &=R_{W\to C}(t_f)\bigl(\mathbf p_{\mathrm{tag}}^W(t_f)-\mathbf c^W(t_f)\bigr),\\
        \xi_f&=\frac{f_xX_f/Z_f+c_x}{w_{\mathrm{img}}},\qquad Z_f>0,
    \end{aligned}
    \label{eq:app:future_relation_projection}
\end{equation}
where $f_x$ and $c_x$ are the horizontal focal length and principal point in pixels, and $w_{\mathrm{img}}$ is the image width in pixels. The coordinate $\xi_f$ is not clipped to the image. For a valid forward projection, the horizontal-band label is
\looseness=-1
\begin{equation}
    B(\xi_f)=\begin{cases}
        \text{out of view, left},&\xi_f<0,\\
        \text{front left},&0\leq\xi_f<0.4,\\
        \text{straight ahead},&0.4\leq\xi_f<0.6,\\
        \text{front right},&0.6\leq\xi_f\leq1,\\
        \text{out of view, right},&\xi_f>1.
    \end{cases}
    \label{eq:app:future_relation_bands}
\end{equation}
The relation uses the future camera orientation and target geometry. Candidates close to the band boundaries are excluded, and the remaining label is mapped to the options in the question. This horizontal-band rule does not assign a label to a point behind the camera or an undefined projection.
\looseness=-1

\section{Metric Geometry and Temporal Fusion}
\label{app:model_implementation}

We supplement the absolute metric depth adapter and spatiotemporal fusion in
Section~\ref{sec:metric_adapter}, using the same depth variables and observation boundary.
The Metric Encoder processes geometry derived from the refined depth $D_{\mathrm m}$,
while the Motion Encoder processes relative camera changes from the multiview
geometry model. Their features are aligned in time and fused into visual tokens
for the progressive state chain.
The $\Omega$ entry in Table~\ref{tab:geometry_ablation} denotes VGGT-$\Omega$~\citep{wang2026vggt}.
The VLM fine-tuning described in Section~\ref{sec:state_chain_learning}
uses LoRA~\citep{hu2021lora}.
\looseness=-1

\paragraph{Global and local depth corrections.}
\label{app:model_depth_adapter}
For $A$ observed geometric anchors, let
$D_0,D_{\mathrm m}\in\mathbb R_{>0}^{A\times H\times W}$ denote the initial and
refined metric depth maps. The global head predicts
$\delta_{\mathrm g}\in\mathbb R^{A\times1\times1}$, which is broadcast across
each frame, and the local head predicts
$\Delta_{\mathrm l}\in\mathbb R^{A\times h\times w}$.
Both heads use multiframe geometric features. The global head aggregates spatial
features to estimate a frame-wide correction; the local head retains spatial
features to estimate nonuniform corrections. The operation $\mathcal U$ in
\Eqref{eq:metric_adapter} bilinearly upsamples each local map to $H\times W$.
Thus the nonnegative coefficients $\gamma_{\mathrm g}$ and $\gamma_{\mathrm l}$ bound dimensionless log-depth
updates, while $D_0$ and $D_{\mathrm m}$ remain in metres. For the update defined
in \Eqref{eq:metric_adapter}, these bounds imply
\looseness=-1
\begin{equation}
    e^{-(\gamma_{\mathrm g}+\gamma_{\mathrm l})}
    \leq \frac{D_{\mathrm m}(a,\mathbf r)}{D_0(a,\mathbf r)}
    \leq e^{\gamma_{\mathrm g}+\gamma_{\mathrm l}},
    \label{eq:app:model_depth_bound}
\end{equation}
where $\mathbf r$ is an image location. The multiplicative form preserves positive
depth and makes the correction proportional to the initial estimate. Its
accuracy consequently depends on the quality of that estimate and of the
geometric features used to refine it.
Complementary reconstruction methods address object motion through
per-timestep pointmaps~\citep{zhang2025monst3r} and motion-aware streaming
state updates~\citep{wang2026raymap3r}. Here, the adapter refines pretrained
depth estimates for metric reasoning.
\looseness=-1

\paragraph{Metric geometry and camera motion.}
\label{app:model_streams}
The refined depth represents distance along the camera's optical axis. At an
anchor with intrinsics $\mathbf K_a$, a pixel $(x,y)$ is lifted to the local
camera frame as
$\mathbf x_a^C=D_{\mathrm m}(a,x,y)\mathbf K_a^{-1}[x,y,1]^\top$.
Metric-surface features combine the depth, camera rays, lifted surface
coordinates, and confidence and validity information. Camera-motion
features from the multiview geometry model describe relative rotations and camera centres in the first valid
anchor's coordinate system. The translation component is normalized by the
accumulated camera path length, with a positive numerical floor, to accommodate
the scale convention of the multiview predictor. Separate spatial encoders
therefore process the metric surface and normalized camera-motion features.
Their outputs at a shared anchor and spatial cell are concatenated and projected
to a common latent width $d$. This preserves the distinct geometric roles of
the two inputs while allowing them to contribute jointly to attention.
\looseness=-1

\paragraph{Anchor-time alignment and fusion.}
\label{app:model_temporal_attention}
RGB frames and geometric anchors retain their source-frame indices within the
observed interval $[i_s,i_o]$, whose endpoints correspond to the observation
boundaries $t_s$ and $t_o$. Their shared normalized temporal coordinate is
$\eta(i)=(i-i_s)/\max(1,i_o-i_s)$; repeated RGB frames retain the same coordinate,
and padded anchors carry an invalid mask. This coordinate represents position
within the input window. Query-time embeddings are added to projected RGB
tokens, and anchor-time embeddings are added to the geometry bank.
\looseness=-1

For attention head $h$, let $\mathbf q_{hq}$ be the visual query at RGB token
$q$, and $\mathbf k_{hk},\mathbf v_{hk}$ the key and value for geometry-bank
token $k$. Each head has width $d_h$. Token $k$ belongs to anchor $a(k)$, giving
$\Delta\eta_{qk}=\eta(i_q)-\eta(i_{a(k)})$. A learned temporal bias
$b_h(\Delta)=\beta_{h1}\Delta+\beta_{h2}|\Delta|
+\beta_{h3}\Delta^2+\beta_{h0}$ augments the content score:
\looseness=-1
\begin{equation}
    \mathbf o_{hq}=\sum_k
    \operatorname{softmax}_k\!\left(
    \frac{\mathbf q_{hq}^{\top}\mathbf k_{hk}}{\sqrt{d_h}}
    +b_h(\Delta\eta_{qk})+M_{qk}\right)\mathbf v_{hk}.
    \label{eq:app:model_attention}
\end{equation}
Here $M_{qk}=0$ for valid geometry tokens and $M_{qk}=-\infty$ for invalid spatial
cells or padded anchors; this validity mask is shared across queries.
After concatenating the head outputs, an output
projection and normalization map the result to the RGB token width:
\looseness=-1
\begin{equation}
    \widetilde{\mathbf x}_q=\mathbf x_q+
    \alpha\,\mathcal P\!\left(\operatorname{Concat}_h\mathbf o_{hq}\right),
    \label{eq:app:model_fusion_residual}
\end{equation}
where $\alpha$ is a learned bounded scalar and $\mathcal P$ includes this
projection and normalization. The resulting visual representation is the fused video representation
$X$ used in Sec.~\ref{sec:state_chain_learning}. Both the RGB queries and geometry
bank are formed from the observed prefix ending at $t_o$; the withheld future
interval supplies supervision targets rather than additional visual inputs.
\looseness=-1

\section{Dataset Composition and Evaluation}
\label{app:integrity}
\subsection{Training and Test Composition}
\label{app:release_integrity}
We use SpatialMind-30K for VLM fine-tuning; it contains 30,677 question--answer pairs comprising 11,456 egocentric and 19,221 driving questions. We evaluate on the SpatialMind-2K benchmark, which contains 2,000 test questions comprising 858 egocentric and 1,142 driving questions. Table~\ref{tab:app:split_family} gives the distribution across the eight task families. The test questions cover 357 source clips from 172 recording groups: 272 egocentric clips from 87 groups and 85 driving segments. A question may refer to a temporal crop, so these counts do not imply one independent recording per question.
\looseness=-1

\begin{table}[!htb]
    \centering
    \caption{Question distribution by source domain and task family. Training statistics describe SpatialMind-30K, and test statistics describe SpatialMind-2K. A zero denotes an absent source-domain and task-family combination.}
    \label{tab:app:split_family}
    \small
    \begin{tabular}{@{}lrrrr@{}}
        \toprule
        & \multicolumn{2}{c}{Training: SpatialMind-30K} & \multicolumn{2}{c}{Test: SpatialMind-2K}\\
        Task family & Ego & Driving & Ego & Driving\\
        \midrule
        L1: Motion state & 2,889 & 2,125 & 143 & 143\\
        L1: Spatial state & 1,968 & 2,312 & 143 & 143\\
        L2: Ego-motion distance & 3,758 & 7,073 & 143 & 143\\
        L2: Target-motion distance & 0 & 1,682 & 0 & 143\\
        L2: Relative motion & 308 & 1,734 & 143 & 143\\
        L3: Ego future direction & 0 & 1,007 & 0 & 143\\
        L3: Target future distance & 1,334 & 1,393 & 143 & 142\\
        L3: Target future relation & 1,199 & 1,895 & 143 & 142\\
        \midrule
        Total & 11,456 & 19,221 & 858 & 1,142\\
        \bottomrule
    \end{tabular}
\end{table}

\subsection{Evaluation Metrics}
\label{app:released_scoring}
We use mean relative accuracy (MRA) for the five numerical task families and accuracy (ACC) for the three multiple-choice task families, following the main-paper protocol. MRA credits numerical predictions at multiple tolerance levels and averages the resulting correctness indicators. It therefore measures closeness to the reference value instead of requiring exact numerical equality. Multiple-choice accuracy compares the selected option with the reference option. Relative motion, ego future direction, and target future relation use ACC; the remaining five families use MRA.
\looseness=-1

For a numerical prediction $\widehat y_i$ and reference $y_i$, let $\epsilon_i(c)$ denote the allowed error at tolerance setting $c$ in the evaluation protocol. Then
\begin{equation}
    \operatorname{MRA}_i=\frac{1}{|\mathcal C|}\sum_{c\in\mathcal C}
    \mathbf1\bigl[|\widehat y_i-y_i|\leq\epsilon_i(c)\bigr],
    \qquad
    \operatorname{ACC}_i=\mathbf1[\widehat a_i=a_i].
    \label{eq:app:task_metrics}
\end{equation}
Here $\mathcal C$ is the set of tolerance settings, and $a_i$ is the reference answer option. Numerical answers use the requested physical unit; a missing or unparseable final answer receives zero credit. External-benchmark results follow each benchmark's original evaluation convention.
\looseness=-1

\section{Additional Analysis}
\label{app:additional_analyses}
\subsection{Persistence as a Dataset Diagnostic}
\label{app:persistence_diagnostic}
We examine how much a future label changes relative to the last observed
state. This analysis uses annotation-derived current states and is therefore
a diagnostic of the question distribution, not an RGB-only predictive
baseline. For target future distance, retaining the current annotated
distance satisfies the item's numerical tolerance on 117 of 143
egocentric questions (81.8\%) and 49 of 142 driving questions (34.5\%).
This distance diagnostic uses the single primary tolerance specified by each
item's benchmark annotation, without averaging over tolerance levels.
For target future relation, retaining the current relation agrees with
the future category on 92 of 143 egocentric questions (64.3\%) and
77 of 142 driving questions (54.2\%). The central current label and the
future label for straight ahead are treated as the same category.
These rates characterize persistence within the selected tasks and
tolerances; they neither establish a bound on model performance nor
measure access to visual information.
\looseness=-1

\begin{figure}[!htb]
    \centering
    \includegraphics[width=5.5in]{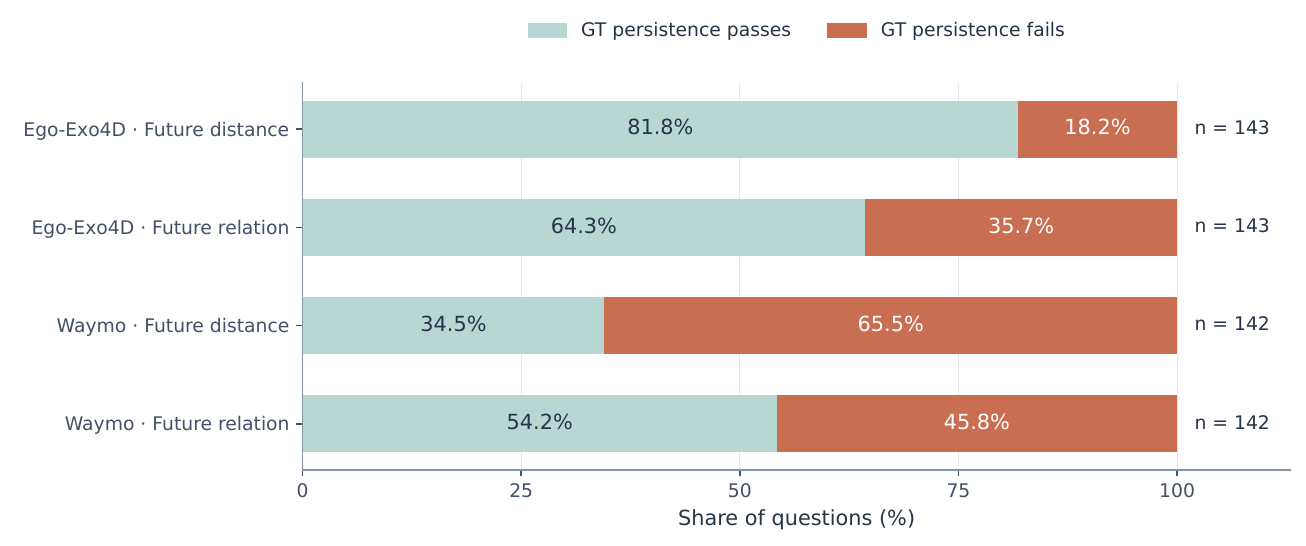}
    \caption{Persistence of annotation-derived current states in future
    distance and relation questions. A distance passes if its unchanged
    current value is within the future item's tolerance; a relation passes
    if the current and future categories match. Counts are 143 per task
    in the egocentric domain and 142 per task in driving. ``GT persistence''
    uses the annotated current state and is not a
    model receiving only RGB.}
    \label{fig:app:future_persistence}
\end{figure}

\subsection{Numerical Error Composition}
\label{app:numeric_error_composition}
Figure~\ref{fig:app:numeric_error_composition} summarizes the recorded
StateChain, Free-form, and Qwen3-VL-8B predictions on all 1,286 numerical
questions in SpatialMind-2K. We normalize each absolute error by the item's
primary tolerance and retain unparseable answers as a separate category.
This diagnostic reports question-level proportions at a single tolerance
per item.
\looseness=-1

StateChain, our progressive state-chain reasoning setting, places
45.26\% of predictions within tolerance, compared with 31.03\% for
Free-form and 12.52\% for Qwen3-VL-8B. The share of errors exceeding
four times the tolerance is 13.61\%, compared with 26.83\% and 57.85\%,
respectively. Relative to Free-form, StateChain reduces the frequency of
these large errors by 49.3\%. The distribution shows an advantage in both
numerical precision and the avoidance of large deviations.
\looseness=-1

\begin{figure}[!htb]
    \centering
    \includegraphics[width=5.5in]{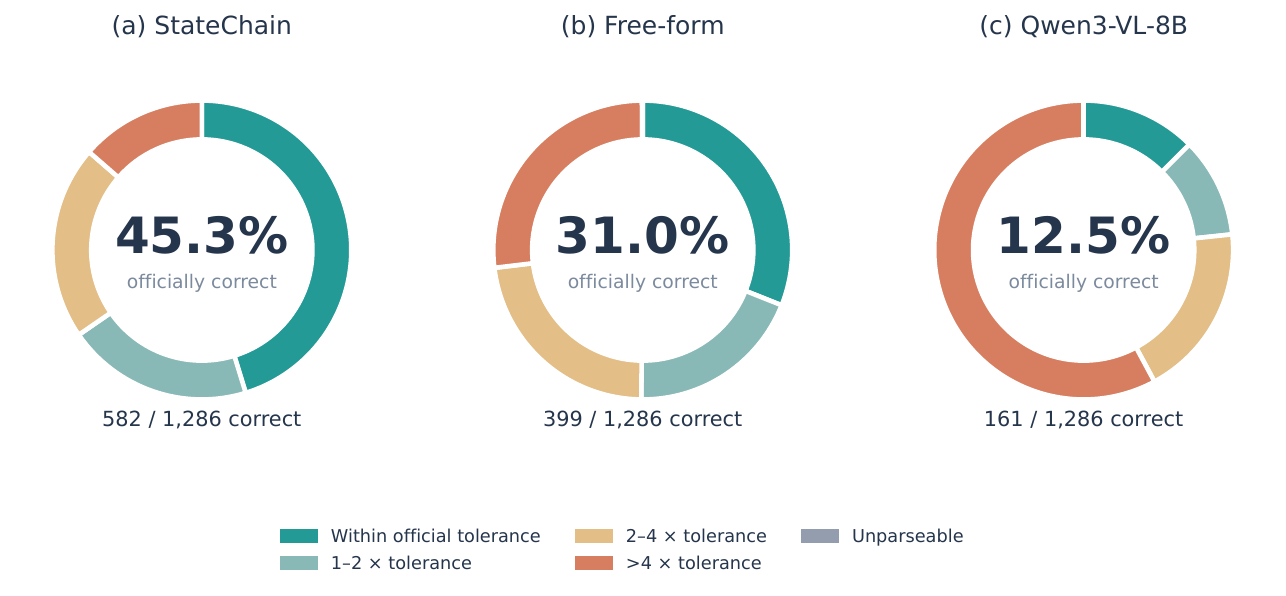}
    \vspace{-0.5em}
    \caption{\textls[-10]{Numerical-error composition on the 1,286 numerical questions in
    SpatialMind-2K. \mbox{StateChain} denotes our progressive state-chain reasoning
    setting. Wedges group absolute error divided by the item's primary tolerance
    into $[0,1]$, $(1,2]$, $(2,4]$, and $(4,\infty)$; unparseable responses
    remain in the denominator as a separate category. Ring centers show the
    within-tolerance percentages, with the corresponding counts below.}\looseness=-1}
    \label{fig:app:numeric_error_composition}
\end{figure}

\section{Relationship to Other Spatiotemporal Tasks}
\label{app:related_scope}

\paragraph{A connected progression from spatial states to prediction.}
SpatialMind organizes eight tasks into current-state understanding,
observed-dynamics understanding, and future prediction. This progression
connects metric spatial estimates to changes over an observed interval and then
to conditions beyond its endpoint. Its future tasks explicitly distinguish the
observation boundary $t_o$ from the queried time $t_f>t_o$. Future reasoning is
also an explicit component of STRIDE-QA~\citep{ishihara2026stride}, whose
evaluation interface includes future distance, bearing, and ego-speed queries.
SpatialMind complements this setting with a common three-level task structure
and progressive state-chain supervision, allowing current estimates, observed
changes, and future targets to be studied within the same framework.
This emphasis on future states differs from question answering about past
observations in OpenEQA's episodic-memory
setting~\citep{majumdar2024openeqa} and from retrieving previously mapped
static geometry across revisits~\citep{wang2026cognimap3d}.
\looseness=-1

\paragraph{Driving and everyday first-person settings.}
SpatialMind combines Waymo driving videos and Ego-Exo4D everyday interactions,
covering both traffic scenes and first-person activities with substantial
changes in viewpoint and spatial context. Questions identify targets through
linguistic descriptions linked to persistent entities. STRIDE-QA's released
interface instead provides visually marked targets in four observed images
at $-1.5$, $-1.0$, $-0.5$, and $0$ seconds. The use of linguistic descriptions
makes target grounding an explicit part of SpatialMind's spatial reasoning task.
SpatialMind-2K's variable observation
windows extend to $19.8$ seconds, supporting longer temporal evidence than the
$1.5$-second span of this published STRIDE-QA interface; these durations describe
observation spans rather than the number of frames sampled by a model.
The two-domain design also supports
examining the same reasoning hierarchy under different scene and motion
conditions; it does not require treating their annotation sources as identical.
\looseness=-1

\paragraph{Traceable geometric targets.}
MLLM-4D~\citep{yin2026mllm4dvisualbasedspatialtemporalintelligence} studies time-evolving object--camera relationships
in three-dimensional space, making it a relevant comparison for reasoning
about dynamic geometry. SpatialMind emphasizes the connection between this
reasoning and explicitly computed supervision: entity identities, coordinate
references, measurement intervals, and geometric operations determine the
reference answers and state-chain targets. These records make it possible to
check whether a distance, path length, direction, or future relation corresponds
to the quantity requested by the question. Together with numerical and
categorical evaluation, this provides a traceable basis for analyzing errors
across the three levels. Comparisons retain each benchmark's input and scoring
conventions; shared spatial terminology alone does not imply identical tasks.
\looseness=-1

\section{Limitations and Future Work}
\label{app:limitations}

SpatialMind depends on estimated geometry, correspondences, and trajectories,
whose errors can propagate to answers and state-chain targets. Its two source
domains and the prediction horizons evaluated in the benchmark do not establish
unrestricted transfer or long-term forecasting. Readable state chains do not
establish causal faithfulness, and the reported experiments do not characterize
end-to-end latency. Future
work will quantify sensitivity to geometric
errors and extend evaluation to longer horizons and more diverse motion,
alongside matched-protocol comparisons and explicit measurements of inference
cost.
For longer observation histories, recurrent geometric
states~\citep{wang2025continuous} and instance anchors for long-sequence
alignment~\citep{gao2026list3r} offer possible representations to investigate.
Their value for future spatial prediction remains to be evaluated.
\looseness=-1

\section{Visualizations}
\label{app:visualizations}
The following examples connect observed video, metric geometry, reference-answer computation, and progressive state-chain supervision. Model inputs contain only the observed video and its question, with no future frames or textual information about future states. Future references shown in prediction examples are used to construct supervision and are withheld from model inputs.
\looseness=-1

\subsection{A Worked Example of Reference-Answer Computation}
\label{app:visual_worked_example}
\noindent\textbf{From a video question to structured supervision.}
Figure~\ref{fig:app:level2_relative_motion} illustrates the reference-answer computation for a Level~2 question, linking the observed video and displayed spatial measurements to the numerical decision and progressive state-chain supervision.
\looseness=-1

\medskip
\noindent\textbf{Question header.}
The silver pickup truck with a black roof rack is the target. The question concerns changes in its distance and direction relative to the observer. Its five options distinguish approaching, moving rightward, remaining approximately unchanged, moving leftward, and receding. This wording determines which geometric relation must be measured throughout the clip.
\looseness=-1

\medskip
\noindent\textbf{A: Video clip.}
The observed window contains 16 consecutive frames sampled at 5~fps, spanning $(16-1)/5=3.0$~s from the first to the last frame. The six displayed thumbnails summarize this window while preserving the target's identity. Bounding boxes help readers locate the pickup; they are not part of the model's video input. The target remains visible throughout the observed window used for the calculation.
\looseness=-1

\medskip
\noindent\textbf{B: Depth and semantics.}
At the last observed frame, metric depth describes the scene's distance structure and the semantic map locates the target within the scene. Together, these views connect the appearance-based referring expression to a spatial entity. The relative-motion calculation uses the target's annotated three-dimensional centres, whose camera-relative ranges and bearings are expanded in the subsequent computation figure.
\looseness=-1

\medskip
\noindent\textbf{C: Geometry and decision.}
The spatial view and computation summary show range increasing from $12.603$ to $18.165$~m, alongside a small rightward bearing change from approximately $0.034^\circ$ to $0.652^\circ$. Applying \Eqref{eq:app:driving_relative_normalization} compares normalized radial change with angular change in radians. Radial motion dominates, and the trend check passes, yielding the receding relation and option~E. A small rightward change therefore does not determine the categorical answer by itself.
\looseness=-1

\medskip
\noindent\textbf{D: Progressive state chain.}
The grounding block fixes the target, observation scope, reference frame, and requested quantity. The current-state block records the endpoint measurements. The dynamics block expresses their signed changes and resulting motion relation, and the answer block maps that relation to E. This structure preserves the connection between measured geometry and the final categorical supervision.
\looseness=-1

\begin{figure}[p]
    \centering
    \includegraphics[page=1,height=0.88\textheight,width=5.5in,keepaspectratio]{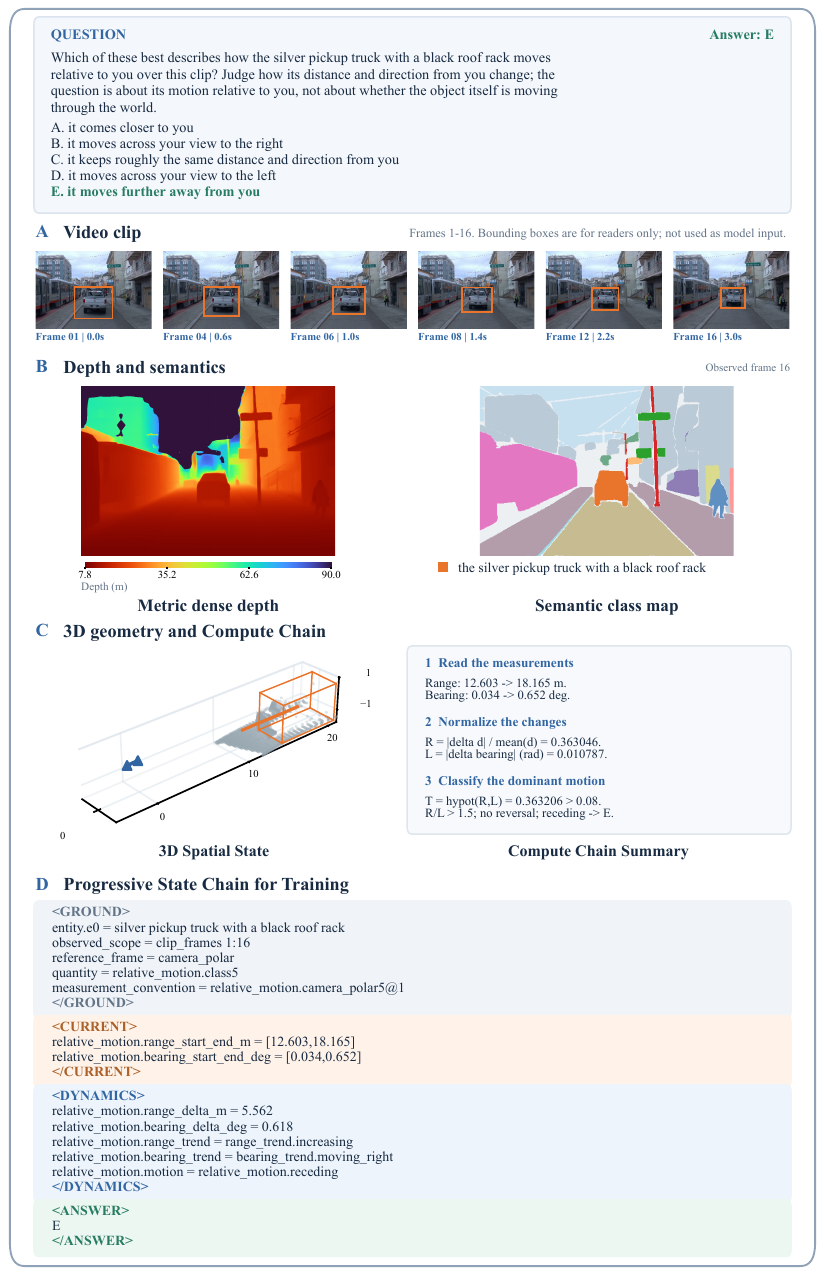}
    \caption{Level~2 relative-motion example, from the question and observed video to structured supervision. The target pickup recedes from the observer, giving option E. The question header and panels A--D show the observed clip, depth and semantics, geometric calculation, and progressive state-chain target.}
    \label{fig:app:level2_relative_motion}
\end{figure}
\clearpage
\subsection{The Full Compute Chain for the Same Example}
\label{app:visual_full_compute_chain}
\noindent\textbf{Expanded reference-answer computation.}
Figure~\ref{fig:app:level2_compute_chain} expands the same example into seven steps, exposing the measurements and checks behind its reference label.
\looseness=-1

\medskip
\noindent\textbf{01: Target grounding.}
The referring expression identifies the pickup. Verification on two observed frames gives intersection-over-union values of $0.514$ and $0.543$, both marked as passing.
\looseness=-1

\medskip
\noindent\textbf{02: Observation window.}
The selected sequence supplies 16 consecutive observations over 3.0~s. All core motion operands come from this window, with the target visible in every frame.
\looseness=-1

\medskip
\noindent\textbf{03: Spatial quantities.}
The table lists optical-frame centres in metres, with axes pointing right, down, and forward. Range is the centre's Euclidean norm, rounded to millimetres. Bearing is the right-positive horizontal angle, displayed as $\beta_k$ in degrees; thus $\alpha_k=\pi\beta_k/180$ in the preceding equations.
\looseness=-1

\medskip
\noindent\textbf{04: Normalization and classification.}
All 16 ranges give $\bar d=15.320375$~m. Their endpoint difference is $5.562$~m, and the bearing change is $0.618024^\circ$. \Eqref{eq:app:driving_relative_normalization} gives $R=0.363045944$, $L=0.010786549$, and $T=0.363206149$. With zero-based midpoint $m=8$, the products $(d_m-d_0)(d_{15}-d_m)=7.726905$~$\mathrm{m}^2$ and $(\beta_m-\beta_0)(\beta_{15}-\beta_m)\simeq0.095344$~$\mathrm{deg}^2$ are positive: the two halves agree in sign. The ordered classifier rules out unchanged motion because $T\geq0.08$, passes these reversal checks, and selects receding because $R\geq1.5L$ and $\Delta d>0$. A dominant-axis sign reversal or lack of dominance would exclude a candidate.
\looseness=-1

\medskip
\noindent\textbf{05: Alternative readings.}
World displacement measures a different quantity: the object and observer displacements are $6.657$ and $1.086$~m. Pixel-size estimates and fitted trends provide cross-checks; agreement here does not validate mixing metres with degrees or reversing the bearing convention. The pixel-derived final range, $20.150$~m, also differs from the annotated $18.165$~m.
\looseness=-1

\medskip
\noindent\textbf{06: Answer mapping.}
Receding maps to E under exact option matching, with $20\%$ uniform chance accuracy. The inset's nine unsuccessful blind rules describe this sample only, not a benchmark-wide evaluation.
\looseness=-1

\medskip
\noindent\textbf{07: Executable verification.}
Recomputation reproduces E, with range/bearing residuals at most $10^{-9}$. The optical-to-camera mapping $(x,y,z)\mapsto(z,-x,-y)$, followed by extrinsics and pose, recovers the world centre with maximum residual $7.11\times10^{-15}$~m. These checks establish agreement between the displayed quantities and their geometric source. The residuals verify numerical and coordinate consistency, not absolute measurement accuracy.
\looseness=-1

\begin{figure}[p]
    \centering
    \includegraphics[page=2,height=0.90\textheight,width=5.5in,keepaspectratio]{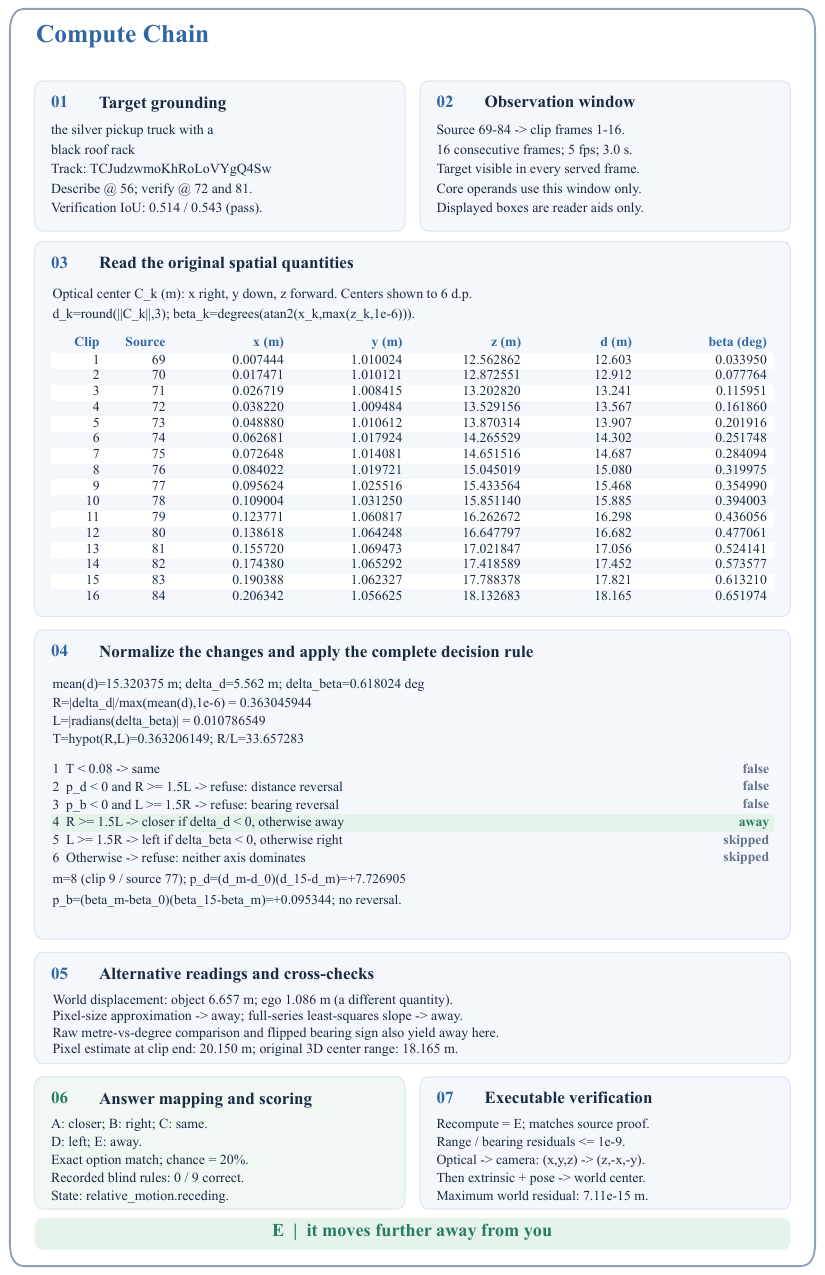}
    \caption{Reference-answer computation for Figure~\ref{fig:app:level2_relative_motion}. This is the expanded record behind the compressed calculation in panel C: grounding and temporal scope, per-frame measurements, normalization and decision rules, cross-checks, answer mapping, and numerical verification. The chain documents how the displayed measurements determine the answer and its supervision.}
    \label{fig:app:level2_compute_chain}
\end{figure}
\clearpage
\begin{figure}[p]
    \centering
    \includegraphics[page=1,height=0.87\textheight,width=5.5in,keepaspectratio]{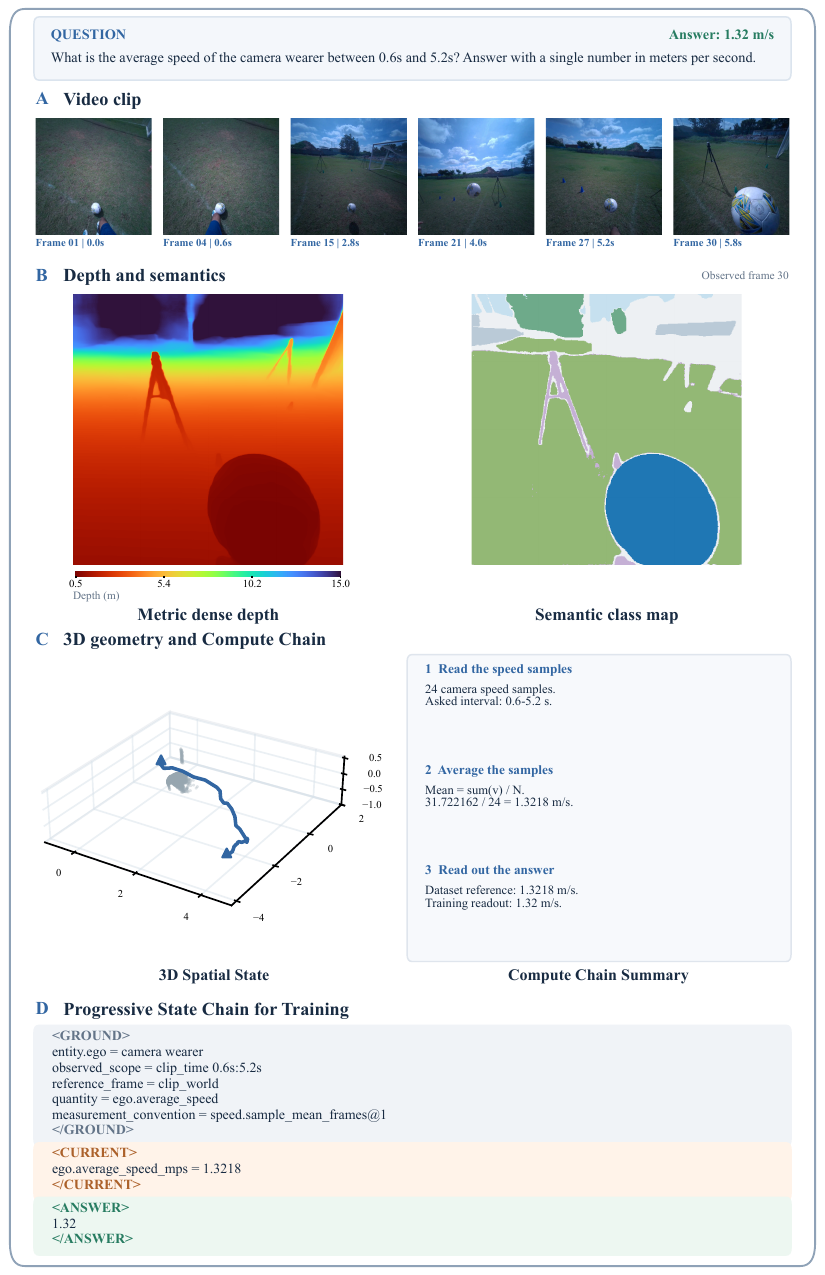}
    \caption{Level~1 motion-state example of egocentric scene: averaging 24 camera-speed samples over $0.6$--$5.2$~s gives $1.3218$~m/s, with $1.32$~m/s as the training answer. The displayed state chain is a supervision target.}
    \label{fig:app:l1_ego_average_speed}
\end{figure}
\clearpage
\begin{figure}[p]
    \centering
    \includegraphics[page=2,height=0.87\textheight,width=5.5in,keepaspectratio]{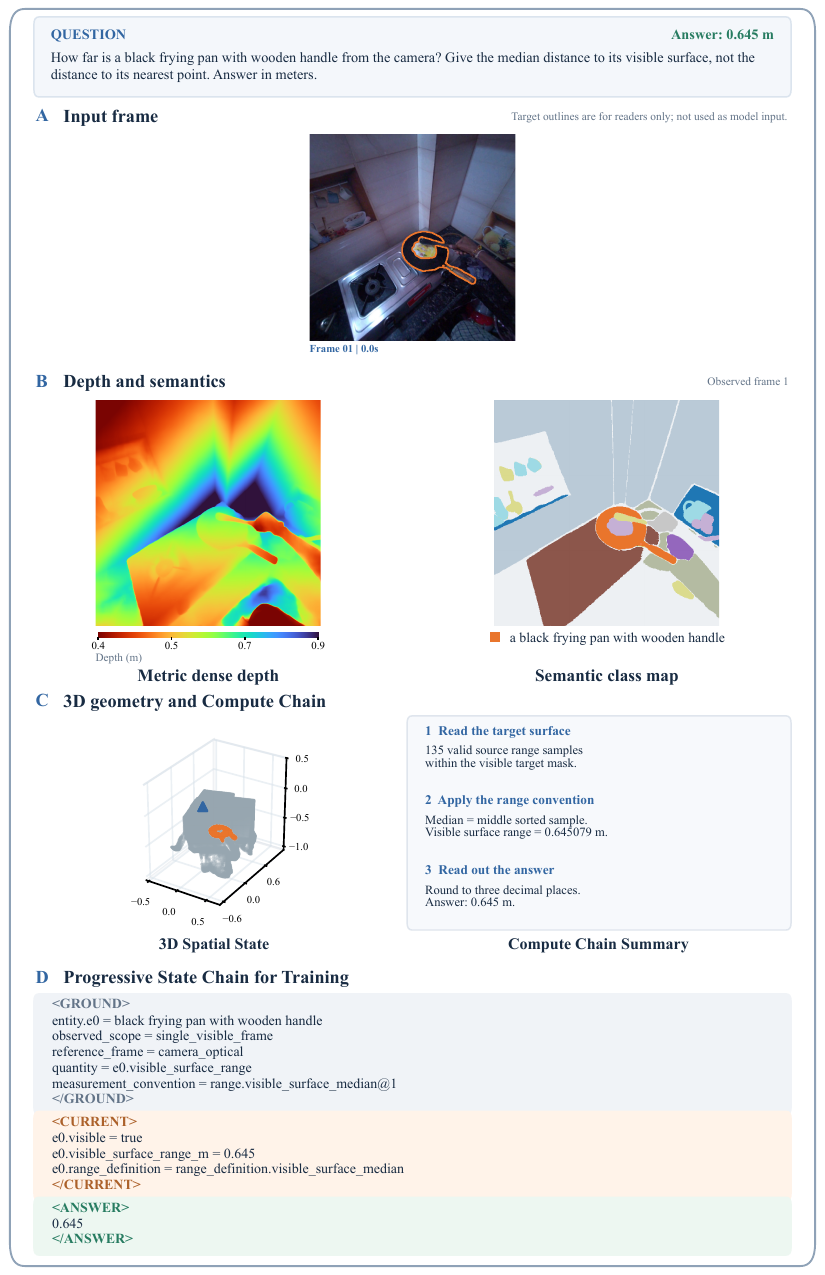}
    \caption{Level~1 spatial-state example of egocentric scene: the median range of 135 valid samples on the frying pan's visible surface is $0.645$~m. This question uses a visible-surface statistic rather than an object-centre or nearest-point distance; outlines are reader annotations, and the state chain is a training target.}
    \label{fig:app:l1_pan_surface_range}
\end{figure}
\clearpage
\begin{figure}[p]
    \centering
    \includegraphics[page=4,height=0.87\textheight,width=5.5in,keepaspectratio]{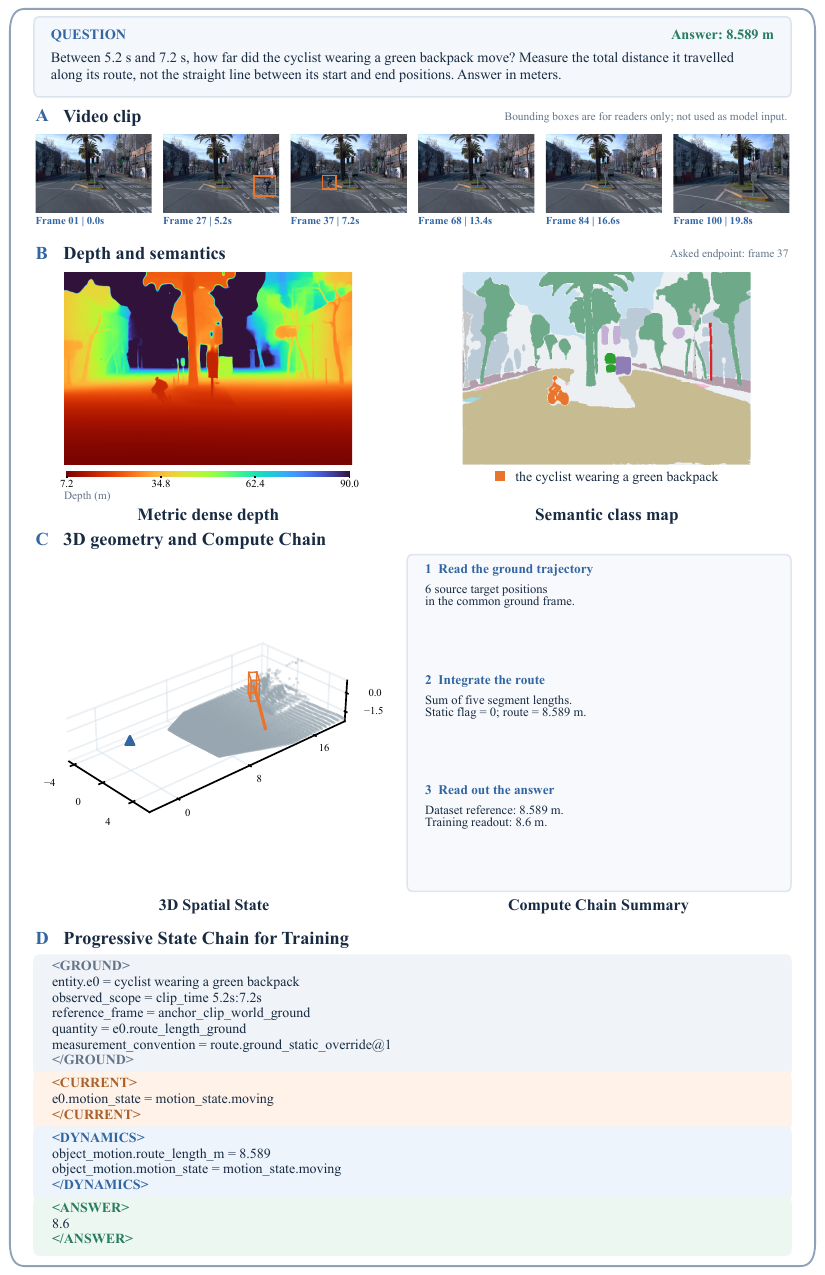}
    \caption{Level~2 target-motion example: summing five ground-trajectory segments over $5.2$--$7.2$~s gives the cyclist a route length of $8.589$~m, with $8.6$~m as the training answer. The route uses six source positions; boxes are reader annotations, and the chain is a supervision target.}
    \label{fig:app:l2_cyclist_route}
\end{figure}
\clearpage
\begin{figure}[p]
    \centering
    \includegraphics[page=6,height=0.87\textheight,width=5.5in,keepaspectratio]{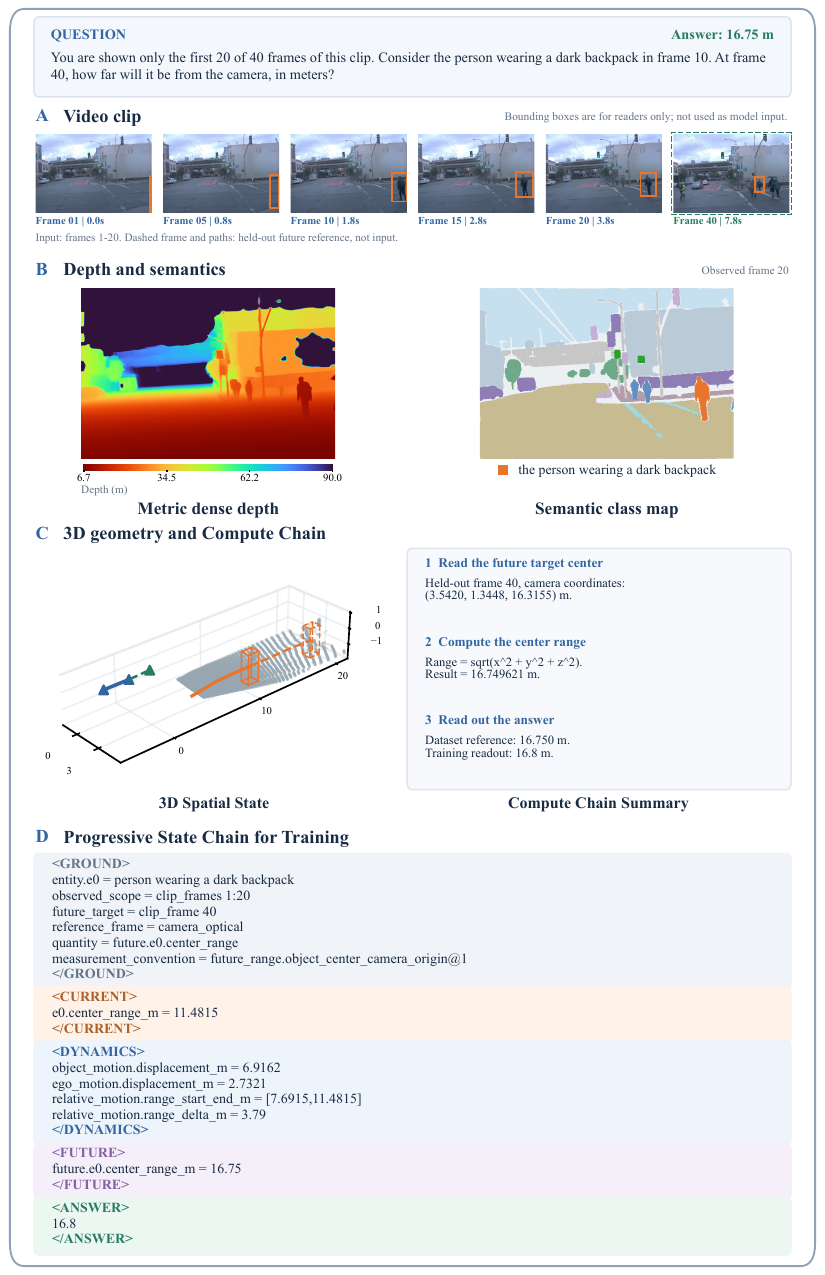}
    \caption{Level~3 target-future-distance example: frames 1--20 are supplied to query the distance from the backpack wearer's centre to the optical camera origin at frame 40. The displayed reference is $16.75$~m and the serialized training answer is $16.8$~m. Dashed future references and chain targets illustrate supervision; neither future frames nor future-state text is supplied at inference.}
    \label{fig:app:l3_target_future_range}
\end{figure}
\clearpage
\endgroup

\end{document}